\documentclass[11pt,reqno,oneside]{amsart}

\usepackage{graphicx,subfigure,threeparttable}
\usepackage{amssymb}
\usepackage{epstopdf}

\usepackage[a4paper, total={6in, 9in}]{geometry}

\usepackage{booktabs} 
\usepackage{subfig} 

\usepackage{amsfonts}

\usepackage{amsmath,amsthm,mathrsfs,amssymb,cite}
\usepackage[usenames]{color}

\theoremstyle{definition}

\theoremstyle{remark}

\numberwithin{equation}{section}

\numberwithin{equation}{section}

\newcounter{saveeqn}

\newcommand{\beq}{\begin{equation}}
\newcommand{\eeq}{\end{equation}}

\title[]{RL-PINNs: Reinforcement Learning-Driven Adaptive Sampling for Efficient Training of PINNs}

\author{Zhenao Song}
\address{School of Mathematics, Southeast University, Nanjing 210096, China.}
\email{220231943@seu.edu.cn}

\date{} 
\begin{document}

\maketitle

\begin{abstract}

Physics-Informed Neural Networks (PINNs) have emerged as a powerful framework for solving partial differential equations (PDEs). However, their performance heavily relies on the strategy used to select training points.   Conventional adaptive sampling methods, such as residual-based refinement, often require multi-round sampling and repeated retraining of PINNs, leading to computational inefficiency due to redundant points and costly gradient computations—particularly in high-dimensional or high-order derivative scenarios.   To address these limitations, we propose RL-PINNs, a reinforcement learning (RL)-driven adaptive sampling framework that enables efficient training with only a single round of sampling.   Our approach formulates adaptive sampling as a Markov decision process, where an RL agent dynamically selects optimal training points by maximizing a long-term utility metric.   Critically, we replace gradient-dependent residual metrics with a computationally efficient function variation as the reward signal, eliminating the overhead of derivative calculations.   Furthermore, we employ a delayed reward mechanism to prioritize long-term training stability over short-term gains.   Extensive experiments across diverse PDE benchmarks, including low-regular, nonlinear,  high-dimensional, and high-order problems, demonstrate that RL-PINNs significantly outperforms existing residual-driven adaptive methods in accuracy.   Notably, RL-PINNs achieve this with negligible sampling overhead, making them scalable to high-dimensional and high-order problems.

\medskip

\noindent{\bf Keywords:}~~ Physics-informed neural networks,  adaptive sampling, reinforcement learning.


\end{abstract}

\section{Introduction}

Partial differential equations (PDEs) are foundational tools for modeling physical phenomena across engineering, physics, and applied mathematics. Traditional numerical methods, such as finite element and finite difference schemes, often struggle with the curse of dimensionality, an exponential escalation in computational cost as problem dimensions increase\cite{32,33}. This limitation has spurred interest in scientific machine learning, where neural networks leverage their high-dimensional approximation capabilities to solve PDEs in a mesh-free manner\cite{34,35}. Among these approaches, Physics-Informed Neural Networks (PINNs)\cite{1,2,3,4,5,6,7} have gained prominence by embedding physical laws directly into neural network training through residual loss terms, enabling solutions that inherently satisfy governing equations.

A critical challenge in PINNs lies in the strategic selection of training points (collocation points). While uniform random sampling is widely adopted, it often fails to resolve regions with sharp gradients, discontinuities, or multiscale dynamics, leading to inefficient convergence and poor solution accuracy\cite{8,9}. To address this, adaptive sampling strategies like Residual-Based Adaptive Refinement (RAR)\cite{12} and Residual-Based Adaptive Distribution (RAD)\cite{16} dynamically allocate more points to regions with high PDE residuals. However, these methods suffer from three key limitations: (1) Multi-round sampling requiring repeated neural network retraining, leading to $3-5\times$ computational overhead; (2) Dependence on exhaustive candidate point evaluation, which becomes prohibitively expensive for high-dimensional or high-order PDEs due to gradient calculations; (3) point selection prioritizes immediate error reduction over long-term training stability, that is some sampling points are redundant.

Recent advances in reinforcement learning (RL) have demonstrated its potential for adaptive decision-making in scientific computing\cite{19,20,21,22,24,25}, particularly in adaptive mesh refinement (AMR)\cite{26,27,28}.  However, RL-driven AMR methods remain constrained by their reliance on mesh-based discretization, limiting scalability for high-dimensional problems.  Building on these insights, we propose RL-PINNs, a novel RL framework that redefines adaptive sampling for PINNs as a sequential decision-making process.  Our key contributions are:

\begin{itemize}
	\item Single-Round Efficient Sampling: By formulating point selection as a Markov decision process (MDP), RL-PINNs eliminates the need for iterative retraining.   An RL agent learns an optimal sampling policy through interactions with the PDE solution space, achieving comprehensive coverage in one training cycle.
	\item Gradient-Free Reward Design: We introduce the function variation as the reward signal, a computationally efficient measure of solution complexity. This replaces traditional gradient-dependent residual metrics, which can be efficiently extended to higher-dimensional and higher-order cases.
    \item Delayed Reward Mechanism: A semi-sparse reward strategy prioritize long-term utility over short-term gains to avoid redundant sampling points.

\end{itemize}

 The rest of the paper is organized as follows. Section 2 provide an overview of PINNs and adaptive sampling strategies. The RL-PINNs is discussed in Section 3.  In Section 4, we use six numerical experiments to verify the efficiency of RL-PINNs. Finally, we give some concluding remarks in Section 5.

\section{Preliminaries}

\subsection{Preliminaries of PINNs}
\ 
\newline

Physics-Informed Neural Networks (PINNs) represent a paradigm shift in solving partial differential equations (PDEs) by combining deep learning with physical constraints. Given a general PDE formulation:

\begin{equation}\label{eq:1}
    \begin{aligned}
\left\{\begin{array}{l}
\mathcal{N}[u](\mathbf{x})=0, \quad \mathbf{x} \in \Omega, \\ 
\mathcal{B}[u](\mathbf{x})=0, \quad \mathbf{x} \in \partial \Omega,
\end{array}\right.
    \end{aligned}
\end{equation}

where $\mathcal{N}$ denotes the differential operator, $\mathcal{B}$ the boundary operator, PINNs approximate the solution $u(\mathbf{x})$ using a neural network $u_{\theta}(\mathbf{x})$  parameterized by $\theta$. The learning objective is formulated through a composite loss function:

\begin{equation}\label{eq:2}
\begin{aligned}
\mathcal{L}(\theta)=\lambda_r \mathcal{L}_r(\theta)+\lambda_b \mathcal{L}_b(\theta),
\end{aligned}
\end{equation}

where

\begin{itemize}
	\item PDE Residual Loss: $\mathcal{L}_r=\frac{1}{N_r} \sum_{i=1}^{N_r}\left|\mathcal{N}\left[u_\theta\right](\mathbf{x}_r^i)\right|^2$ enforces the governing equation over collocation points $\left\{\mathbf{x}_r^i\right\}_{i=1}^{N_r}$.
	\item Boundary Condition Loss: 
$\mathcal{L}_b=\frac{1}{N_b} \sum_{j=1}^{N_b}\left|\mathcal{B}\left[u_\theta\right](\mathbf{x}_b^j)\right|^2$ ensures compliance with spatial-temporal boundaries.
\end{itemize}

The weighting coefficients $\lambda_r$, $\lambda_b$ balance the contribution of each constraint. Training proceeds via gradient-based optimization(e.g., Adam\cite{29}, L-BFGS\cite{30}) to minimize $\mathcal{L}(\theta)$, effectively solving the PDE in a mesh-free manner.

While PINNs circumvent mesh generation and demonstrate universal approximation capabilities, their practical performance critically depends on the selection of collocation points.  Uniform or random sampling often leads to excessive sampling in smooth regions and under-resolution of critical features (e.g., sharp gradients or discontinuities), necessitating adaptive strategies for optimal point allocation.

\subsection{Preliminaries of adaptive sampling}
\ 
\newline

Adaptive sampling aims to strategically select collocation points during PINN training to improve solution accuracy and convergence. Residual-driven adaptive sampling methods dynamically allocate more points to regions where the PDE residual error is large.  Two representative approaches include:

\begin{itemize}
	\item Residual-Based Adaptive Refinement (RAR): Iteratively enriches the training set by adding points where the residual $\mathcal{N}\left[u_\theta\right](\mathbf{x})$ exceeds a threshold. At each iteration, candidate points are evaluated on a pre-defined grid or random subset, and the $top-k$ highest-residual points are appended to the collocation set. While effective for low-dimensional problems, RAR scales poorly with dimensionality due to its reliance on exhaustive candidate evaluations (Algorithm \uppercase\expandafter{\romannumeral1}).

\begin{table}[htbp]
\begin{tabular}{p{1.5 cm} p{11cm}}
\toprule
\multicolumn{2}{l}{\textbf{Algorithm \uppercase\expandafter{\romannumeral1}: RAR}}\\
\midrule
{\bf Require}:& Neural network $u_{\theta}(\mathbf{x})$, initial collocation points $\mathcal{N}_r$, boundary points $\mathcal{N}_b$. Maximum iteration count $T_{max}$.\\
\quad \quad 1.& $t \leftarrow 1$.\\
\quad \quad 2.& Pre-train $u_{\theta}(\mathbf{x})$ for a certain number of iterations.\\
\quad \quad 3.& {\bf while} $ t \leq T_{max}$ {\bf do}\\
\quad \quad 4.& \quad \quad  Sample candidate points $\mathcal{S}_0$ from uniform distribution.\\
\quad \quad 5.& \quad \quad  Compute the PDE residuals for the points in $\mathcal{S}_0$. \\
\quad \quad 6.& \quad \quad  $\mathcal{S} \leftarrow m$ points with the largest residuals in $\mathcal{S}_0$.\\
\quad \quad 7.& \quad \quad $\mathcal{N}_r \leftarrow \mathcal{N}_r \cup \mathcal{S}$.\\
\quad \quad 8.& \quad \quad Train $u_{\theta}(\mathbf{x})$  for a certain number of iterations.\\
\quad \quad 9.& \quad \quad $t = t + 1$.\\
\quad \quad 10.& {\bf end while}.\\
\bottomrule
\end{tabular}
\end{table}

	\item Residual-Based Adaptive Distribution (RAD):  RAD constructs a probability density function $p(\mathbf{x}) \propto \frac{\mathcal{N}\left[u_\theta\right](\mathbf{x})}{\mathbb{E}\left[\mathcal{N}\left[u_\theta\right](\mathbf{x})\right]}+1$. Then define a probability mass function $\tilde{p}(\mathbf{x})=\frac{p(\mathbf{x})}{\sum_{\mathbf{x} \in \mathcal{S}_0} p(\mathbf{x})}$, and samples collocation points from $\tilde{p}(\mathbf{x})$. This prioritizes high-error regions while maintaining stochasticity. This prioritizes high-error regions while maintaining stochasticity.  However, frequent residual evaluations over large candidate pools incur prohibitive computational costs for high-order derivatives (Algorithm \uppercase\expandafter{\romannumeral2}).

\begin{table}[htbp]
\begin{tabular}{p{1.5 cm} p{11cm}}
\toprule
\multicolumn{2}{l}{\textbf{Algorithm \uppercase\expandafter{\romannumeral2}: RAD}}\\
\midrule
{\bf Require}:& Neural network $u_{\theta}(\mathbf{x})$, initial collocation points $\mathcal{N}_r$, boundary points $\mathcal{N}_b$. Maximum iteration count $T_{max}$.\\
\quad \quad 1.& $t \leftarrow 1$.\\
\quad \quad 2.& Pre-train $u_{\theta}(\mathbf{x})$  for a certain number of iterations.\\
\quad \quad 3.& {\bf while} $ t \leq T_{max}$ {\bf do}\\
\quad \quad 4.& \quad \quad  Sample candidate points $\mathcal{S}_0$ from uniform distribution.\\
\quad \quad 5.& \quad \quad  Compute ${p}(\mathbf{x})$ for the points in $\mathcal{S}_0$. \\
\quad \quad 6.& \quad \quad  $\mathcal{S} \leftarrow m$ points according to $\tilde{p}(\mathbf{x})$.\\
\quad \quad 7.& \quad \quad $\mathcal{N}_r \leftarrow \mathcal{N}_r \cup \mathcal{S}$.\\
\quad \quad 8.& \quad \quad Train $u_{\theta}(\mathbf{x})$  for a certain number of iterations.\\
\quad \quad 9.& \quad \quad $t = t + 1$.\\
\quad \quad 10.& {\bf end while}.\\
\bottomrule
\end{tabular}
\end{table}

\end{itemize}

Despite their utility, residual-driven adaptive sampling strategies face three critical challenges: (1) Multi-Round Training Overhead: Both RAR and RAD require repeated sampling and PINN retraining (typically 3–5 rounds),  increasing total training time by $3-5\times$. (2) High Computational Cost: Residual evaluations involve computing gradients of $\mathcal{N}\left[u_\theta\right]$  via automatic differentiation, which dominates training time for high-dimensional PDEs. (3) Short-Term Optimization Bias: Greedy point selection based on instantaneous residuals neglects long-term training dynamics, often oversampling local features at the expense of global solution stability.

These limitations motivate the need for a single-round, gradient-free adaptive sampling framework that optimizes collocation point selection through long-term planning.

\section{Reinforcement Learning-Driven Adaptive Sampling for PINNs}

This section presents the RL-PINNs framework, which integrates reinforcement learning (RL) into adaptive sampling for PINNs. We formulate the sampling process as a Markov Decision Process (MDP) and employ Deep Q-Networks (DQN)\cite{31} to learn an optimal policy for collocation point selection. The key components of our approach are detailed below.

\subsection{Markov Decision Process Formulation}
\ 
\newline

The adaptive sampling task is modeled as an MDP with the following components:

\begin{itemize}
	\item State Space: The state $s^{(t)} \in \mathcal{S}$ is defined as the current spatial position $\mathbf{x}^{(t)} \in \Omega$, where $\Omega \subseteq$ $\mathbb{R}^d$ is the problem domain.
	\item Action Space: The agent selects discrete actions $a^{(t)} \in \mathcal{A}$, representing incremental displacements along each spatial dimension. For example, in a 2D domain, $\mathcal{A}=$ $\{( \pm \Delta x, 0),(0, \pm \Delta y)\}$, where $\Delta x, \Delta y$ are predefined step sizes.
    \item Reward Function: A delayed semi-sparse reward mechanism is designed to prioritize regions with high solution complexity. The reward is defined as:

\begin{equation}\label{eq:7}
    \begin{aligned}
R^{(t)}= \begin{cases}\delta u^{(t)}, & \text { if } \delta u^{(t)} \geq \varepsilon, \\ 0, & \text { otherwise }\end{cases}
    \end{aligned}
\end{equation}

where $\delta u^{(t)}=\left|u_\theta\left(\mathbf{x}^{(t+1)}\right)-u_\theta\left(\mathbf{x}^{(t)}\right)\right|$ quantifies the function variation between consecutive states, and  is $\varepsilon$ a predefined threshold.

    \item Transition Dynamics: The next state $\mathbf{x}^{(t+1)}$ is deterministically updated as $\mathbf{x}^{(t+1)}=\mathbf{x}^{(t)}+a^{(t)}$.
\end{itemize}

\subsection{DQN Architecture and Training}
\ 
\newline

We implement the DQN framework with the following components:

\begin{itemize}
	\item Q-Network: A neural network $Q(s,a;\eta)$ approximates the action-value function, where $\eta$ denotes trainable parameters. The network takes the state $\mathbf{x}^{(t)}$ as input and outputs Q-values for all possible actions.
	\item Target Network: A separate target network $Q_{tar}\left(s, a ; \eta^{-}\right)$, , with parameters $\eta^{-}$ periodically synchronized from the Q-network, stabilizes training.
    \item Experience Replay: Transitions $\left(\mathbf{x}^{(t)}, a^{(t)}, R^{(t)}, \mathbf{x}^{(t+1)}\right)$ are stored in a replay buffer $\mathcal{P}$. During training, mini-batches are sampled from $\mathcal{P}$ to break temporal correlations.
\end{itemize}

The Q-network is trained by minimizing the mean squared Bellman error:

\begin{equation}\label{eq:9}
    \begin{aligned}
\mathcal{L}(\eta)=\mathbb{E}_{\left(\mathbf{x}, a, R, \mathbf{x}^{\prime}\right) \sim \mathcal{P}}\left[\left(R+\gamma \max _{a^{\prime}} Q_{tar}\left(\mathbf{x}^{\prime}, a^{\prime} ; \eta^{-}\right)-Q(\mathbf{x}, a ; \eta)\right)^2\right],
    \end{aligned}
\end{equation}

where $\gamma \in[0,1)$  is the discount factor.

\subsection{Algorithmic Workflow}
\ 
\newline

The sampling process proceeds as follows:

\begin{itemize}
	\item Initialization: Initialize the Q-network and target network with identical parameters.
	\item Episode Execution: For each episode, the agent starts at a random initial state $\mathbf{x}^{(0)}$ and interacts with the environment for $T$ steps. At each step, an action is selected using an greedy policy to balance exploration and exploitation.
    \item Reward Calculation: The reward $R^{(t)}$ is computed based on the function variation $\delta u^{(t)}$.
    \item Experience Storage: Transitions are stored in $\mathcal{P}$.
    \item Network Update: The Q-network is updated periodically using mini-batches from $\mathcal{P}$, and the target network is synchronized every $K$ episodes($K=5$ in this study).
    \item Sampling Termination: The process terminates when the proportion of high-variation points (with $\delta u \geq \varepsilon$ ) exceeds $50 \%$ for $k$ consecutive episodes($k=5$ in this study).
\end{itemize}

Unlike residual-based adaptive sampling, which requires costly gradient computations for residual evaluation, RL-PINNs use the function variation, a gradient-free metric as the reward signal.  This eliminates the computational overhead of automatic differentiation, particularly beneficial for high-order PDEs.  Additionally, the delayed reward mechanism ensures long-term planning by filtering out redundant points with insignificant solution variations. The complete workflow is outlined in Algorithm \uppercase\expandafter{\romannumeral3}.

\begin{table}[htbp]
\begin{tabular}{p{1.5 cm} p{11cm}}
\toprule
\multicolumn{2}{l}{\textbf{Algorithm \uppercase\expandafter{\romannumeral3}: DQN-Driven Adaptive Sampling for PINNs}}\\
\midrule
{\bf Require}:& Q-network $Q(s,a;\eta)$, target network $Q_{tar}\left(s, a ; \eta^{-}\right)$, replay buffer $P$, discount factor $\gamma$, function variation threshold $\varepsilon$, consecutive episode $k$.\\
\quad \quad 1.& $k \leftarrow 0$.\\
\quad \quad 2.& Initialize $Q$ and $Q_{tar}$ with weights $\eta = \eta^{-}$.\\
\quad \quad 3.& {\bf for} episode $ n = 1:N$ {\bf do}\\
\quad \quad 4.& \quad \quad  Sample initial state $\mathbf{x}^{(0)}$.\\
\quad \quad 5.& \quad \quad  {\bf for} step $ t = 1:T$ {\bf do}\\
\quad \quad 6.& \quad \quad \quad \quad Select action $a^{(t)}$ via greedy policy:$$
a^{(t)}= \begin{cases}\arg \max _a Q\left(\mathbf{x}^{(t)}, a ; \eta\right), & \text { with probability } 1-p, \\ \text { random action } a \in \mathcal{A}, & \text { with probability } p = 0.5/n, \end{cases}
$$\\
\quad \quad 7.& \quad \quad \quad \quad Update state: $\mathbf{x}^{(t+1)}=\mathbf{x}^{(t)}+a^{(t)}$.\\
\quad \quad 8.& \quad \quad \quad \quad Compute reward $R^{(t)}$ using Eq.\eqref{eq:7}.\\
\quad \quad 9.& \quad \quad \quad \quad Store transition $\left(\mathbf{x}^{(t)}, a^{(t)}, R^{(t)}, \mathbf{x}^{(t+1)}\right)$ in $\mathcal{P}$.\\
\quad \quad 10.& \quad \quad  {\bf end for}\\
\quad \quad 11.& \quad \quad Proportion of high-variation points  in the rounds $$
r = \frac {Number\{\mathbf{x} \in \{\mathbf{x}^{(0)},\mathbf{x}^{(1)},...,\mathbf{x}^{(T)}\}  \mid \delta u \geq \varepsilon\}}{T},$$\\
\quad \quad 12.& \quad \quad  {\bf if} $r \geq 50\%$ {\bf then} \\
\quad \quad 13.& \quad \quad \quad \quad  $k \leftarrow k+1$\\
\quad \quad 14.& \quad \quad  {\bf else} \\
\quad \quad 15.& \quad \quad \quad \quad  $k \leftarrow 0$\\
\quad \quad 16.& \quad \quad  {\bf end if} \\
\quad \quad 17.& \quad \quad Update  $\eta$ with Eq.\eqref{eq:9}. Remark: Only one gradient descent.\\
\quad \quad 18.& \quad \quad $\eta^{-} \leftarrow \eta$ every 5 episodes.\\
\quad \quad 19.& \quad \quad {\bf if} $k \geq 5$ {\bf then} \\
\quad \quad 20.& \quad \quad \quad \quad {\bf break} \\
\quad \quad 21.& \quad \quad {\bf end if} \\
\quad \quad 22.& {\bf end for}.\\
\quad \quad 23.& {\bf return} High-variation points $\{\mathbf{x} \in \mathcal{P} \mid \delta u \geq \varepsilon\}$.\\
\bottomrule
\end{tabular}
\end{table}

This framework ensures efficient single-round sampling by leveraging RL’s sequential decision-making capability, significantly reducing computational overhead compared to multi-round residual-driven methods. The proposed RL-PINNs framework operates in three phases: (1) pre-training to initialize the PINN solution, (2) RL-driven adaptive sampling to identify critical collocation points, and (3) final PINN training using the optimized point set. See Algorithm \uppercase\expandafter{\romannumeral4}. 

\begin{table}[htbp]
\begin{tabular}{p{1.5 cm} p{11cm}}
\toprule
\multicolumn{2}{l}{\textbf{Algorithm \uppercase\expandafter{\romannumeral4}: RL-PINNs}}\\
\midrule
{\bf Require}:& Neural network $u_{\theta}(\mathbf{x})$, initial collocation points $\mathcal{N}_r$, boundary points $\mathcal{N}_b$.\\
\quad \quad 1.& Pre-train $u_{\theta}(\mathbf{x})$  for a certain number of iterations.\\
\quad \quad 2.& Sample new collocation poingts $S$ with Algorithm \uppercase\expandafter{\romannumeral3}.\\
\quad \quad 3.& $\mathcal{N}_r \leftarrow \mathcal{N}_r \cup \mathcal{S}$.\\
\quad \quad 4.& Train $u_{\theta}(\mathbf{x})$  for a certain number of iterations.\\
\bottomrule
\end{tabular}
\end{table}

\section{Numerical experiments}

This section presents comprehensive numerical experiments to validate the efficacy of RL-PINNs across diverse PDE benchmarks. We compare RL-PINNs against three baseline methods(UNIFORM, RAR, and RAD) in terms of computational accuracy, efficiency, and scalability. Below, we detail the experimental setup, followed by case-specific results and analysis.

\subsection{Experiment Setup}
\ 
\newline

To ensure a fair comparison, all methods adopt identical network architectures, optimization protocols, and evaluation metrics.  The experimental configuration is as follows:

\subsubsection{Network Architectures}
\ 
\newline

\begin{itemize}
	\item PINNs($u_{\theta}$): A fully connected neural network with seven hidden layers of sizes $[64,128,256,512,256,128,64]$, using Tanh activation function.
	\item DQN($Q(s,a;\eta)$): A shallow network with two hidden layers of sizes $[128,64]$, using ReLU activation.
\end{itemize}

\subsubsection{Optimization Protocol}
\ 
\newline

\begin{itemize}
	\item Optimizers: Adam for training across all cases, except for the Burgers’ equation (Case 3), where L-BFGS is additionally applied for fine-tuning.
    \item Pretraining:  (1)$\mathcal{N}_r^{(0)}$: Number of initial collocation points. (2)$iterations^{(0)}$: Number of pretraining iterations. (3)$T_{max}$: Maximum adaptive sampling rounds for RAR and RAD. (4)$\mathcal{S}_0$: Size of the candidate pool evaluated per sampling round. (5)$\mathcal{S}$: Number of points added to the training set per round.

\begin{table}[h]
\centering
\begin{tabular}{ccccccc}  
\toprule
Case & Learning rate & $\mathcal{N}_r^{(0)}$ & $iterations^{(0)}$ & $T_{max}$ & $\mathcal{S}_0$ & $\mathcal{S}$  \\  
\midrule
Single-Peak & 1e-4  & 5000 &5000 & 5 & 1000 & 200 \\  
Dual-Peak & 1e-4  & 5000 &5000 & 5 & 2000 & 400 \\    
Burgers' & 1e-3  & 5000 &5000 & 5 & 1000 & 200 \\
Wave & 1e-3  & 10000 &10000 & 5 & 2000 & 400 \\
High-Dimension & 1e-3  & 10000 &10000 & 5 & 5000 & 1000 \\
High-Order & 5e-5  & 2000 &5000 & 5 & 1000 & 200 \\
\bottomrule
\end{tabular}
\end{table}

	\item The Q-network $Q(s, a ; \eta)$: (1)Learning rate: 1e-3. (2)Discount factor:  $\gamma=0.95$. (3)Action space ($\mathcal{A}$): Discrete displacements along spatial coordinates. (4)Replay buffer ($\mathcal{P}$): Capacity varies per case.

\begin{table}[h]
\centering
\begin{tabular}{ccccccc}  
\toprule
Case & $\mathbf{x}^{(0)}$ & $N$ & $T$ & $\varepsilon$ & $\mathcal{A}$ & $\mathcal{P}$\\  
\midrule
Single-Peak & $x,y \sim U(-0.1,0.1)$ & 100 & 200 &0.005 & $\pm0.1$ &1000  \\  
Dual-Peak & $x,y \sim U(-0.1,0.1)$ & 100 & 400 &0.01 & $\pm0.1$ &2000 \\  
Burgers' & $x \sim U(-0.1,0.1), t \sim U(0,0.1)$ & 100 & 200 &0.1 & $\pm0.1$ &1000 \\
Wave & $x \sim U(-0.5,0.5), t \sim U(0,0.5)$ & 100 & 400 &0.05 & $\pm0.2$ &2000 \\
High-Dimension & $\mathbf{x}_i \sim U(0.4,0.6)$ & 100 & 1000 &0.0001 & $\pm0.1$ &5000 \\
High-Order & $x,y \sim U(0.4,0.6)$ & 100 & 200 &0.05 & $\pm0.1$ &1000 \\
\bottomrule
\end{tabular}
\end{table}

\end{itemize}

\subsubsection{Evaluation Metric}
\ 
\newline

Solution accuracy is quantified using the relative $L_2$ error:

\begin{equation}\label{eq:10}
    \begin{aligned}
\operatorname{err}_{L_2}=\frac{\sqrt{\sum_{i=1}^N\left|\hat{u}\left(\mathbf{x}_i\right)-u\left(\mathbf{x}_i\right)\right|^2}}{\sqrt{\sum_{i=1}^N\left|u\left(\mathbf{x}_i\right)\right|^2}},
    \end{aligned}
\end{equation}

where $N$ denotes the number of test points, and $\hat{u}\left(\mathbf{x}_i\right)$ and $u\left(\mathbf{x}_i\right)$ represent the predicted and exact solutions, respectively.

\begin{figure}[h]
    \centering
{
    \includegraphics[width=0.4\textwidth]{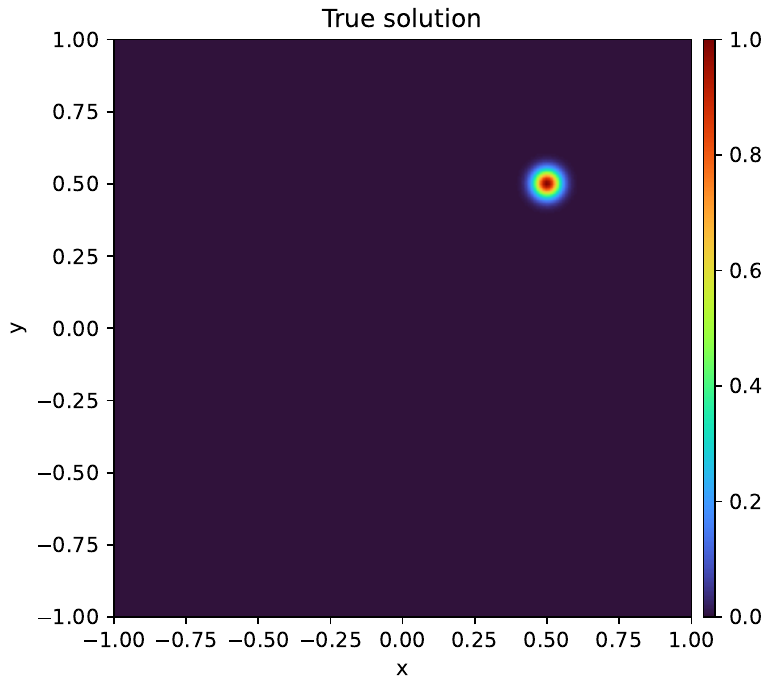}
}
    \caption{\label{fig:peak_true_u} Single-Peak: The exact solution.}
\end{figure}

\subsubsection{Hardware and Software}
\ 
\newline

All experiments are conducted on an NVIDIA RTX 4090  GPU with 64GB memory on a cloud server(https://matgo.cn/host-market/gpu). Implementations utilize PyTorch 2.4.0 for neural network training and Deep Q-Learning.

\subsection{Two-dimensional Poisson equation}
\ 
\newline

This section evaluates RL-PINNs on two variants of the two-dimensional Poisson equation: a single-peak case and a dual-peak case.  Both benchmarks demonstrate the framework’s capability to resolve localized features while maintaining computational efficiency.

\subsubsection{Single-Peak Case}
\ 
\newline

The Poisson equation is defined as:

\begin{figure}[h]
    \centering
    \subfigure[UNIFORM]
{
    \includegraphics[width=0.35\textwidth]{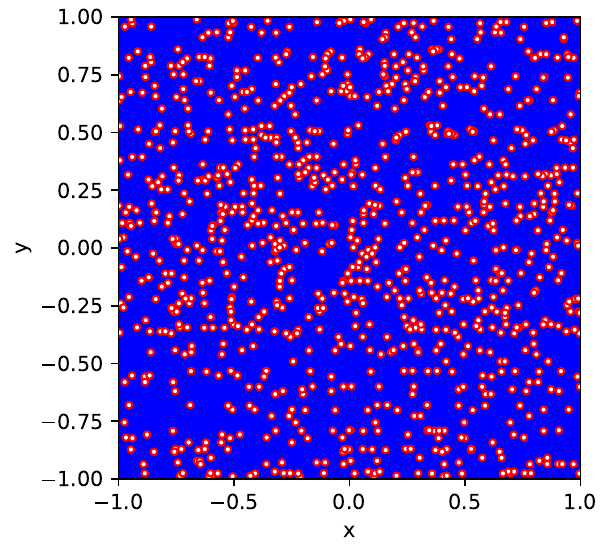}
}
    \subfigure[RAR]
{
    \includegraphics[width=0.35\textwidth]{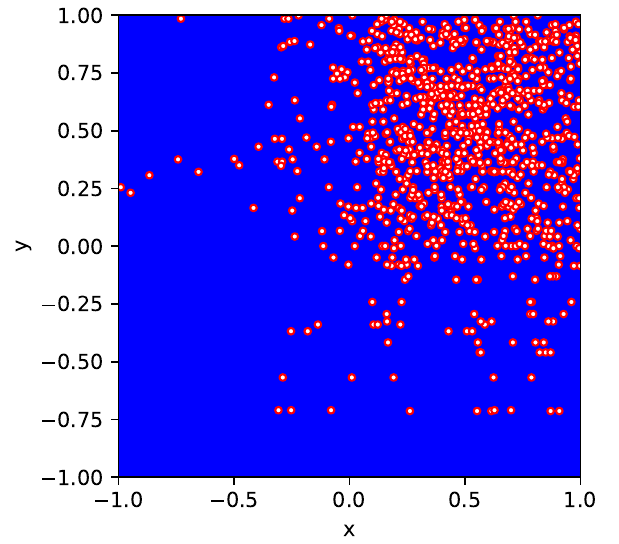}
}
    \subfigure[RAD]
{
    \includegraphics[width=0.35\textwidth]{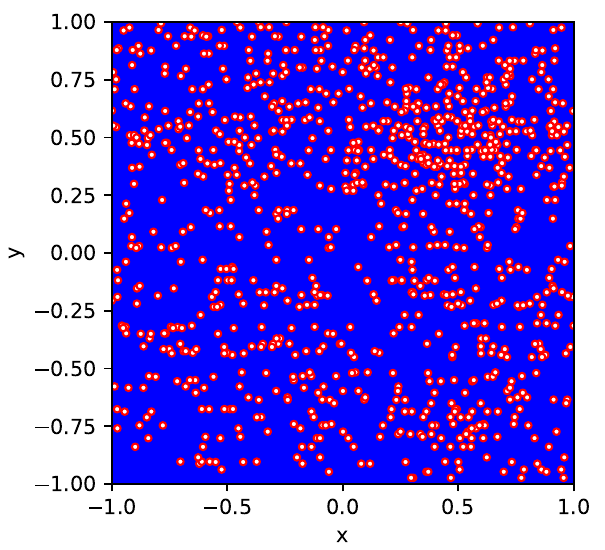}
}
    \subfigure[RL-PINNs]
{
    \includegraphics[width=0.35\textwidth]{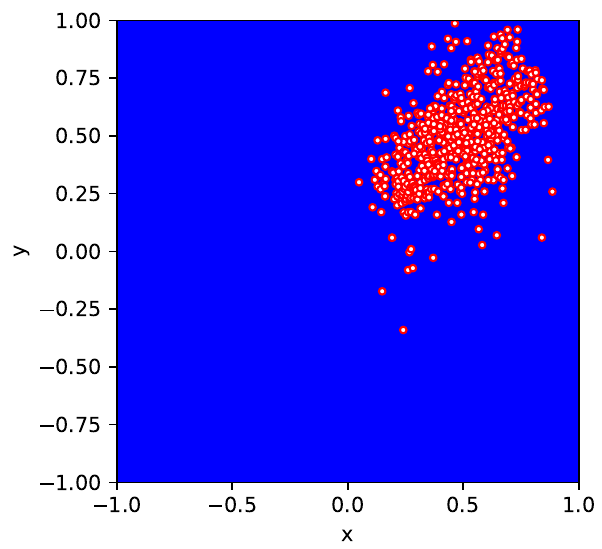}
}
    \caption{\label{fig:peak_points} Single-Peak: Cumulative sampling of UNIFORM, RAR, RAD and one-time sampling of RL-PINNs.}
\end{figure}

\begin{figure}[h]
    \centering
{
    \includegraphics[width=0.2\textwidth]{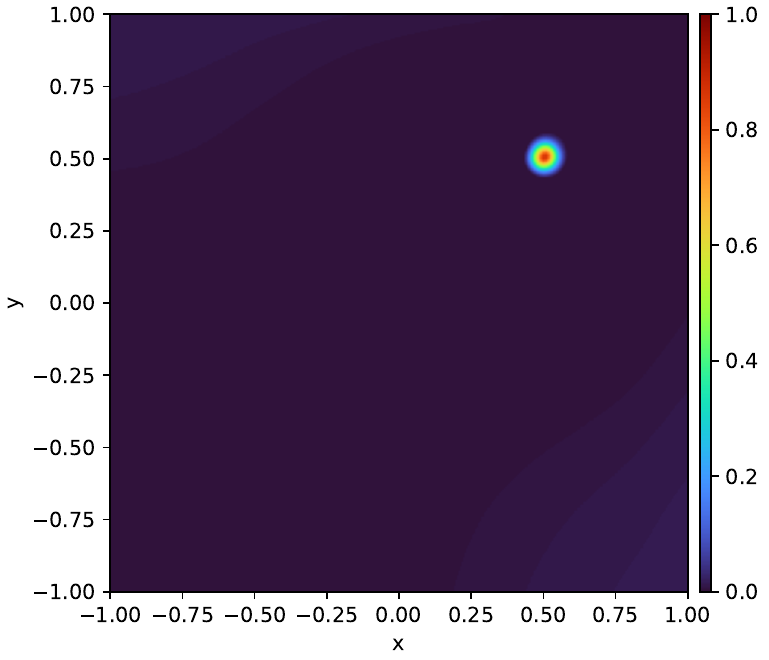}
}
{
    \includegraphics[width=0.2\textwidth]{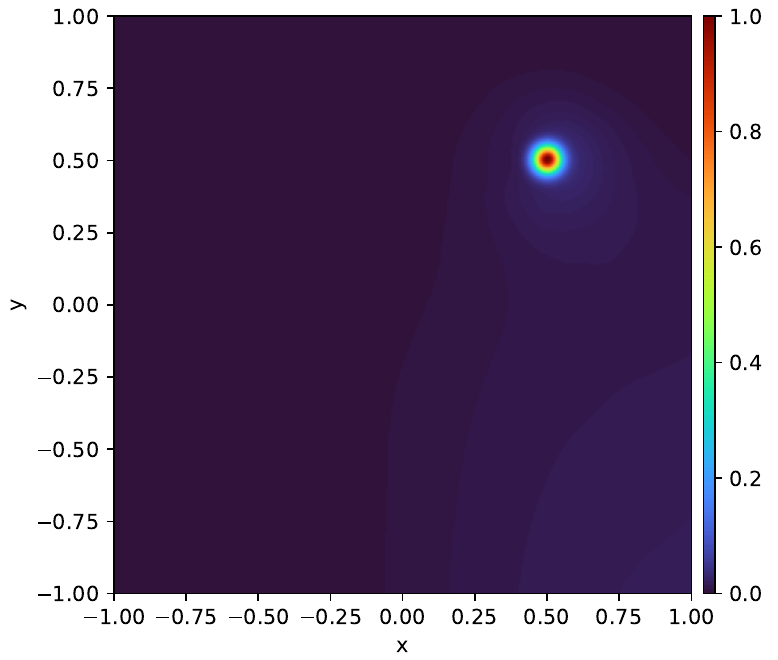}
}
{
    \includegraphics[width=0.2\textwidth]{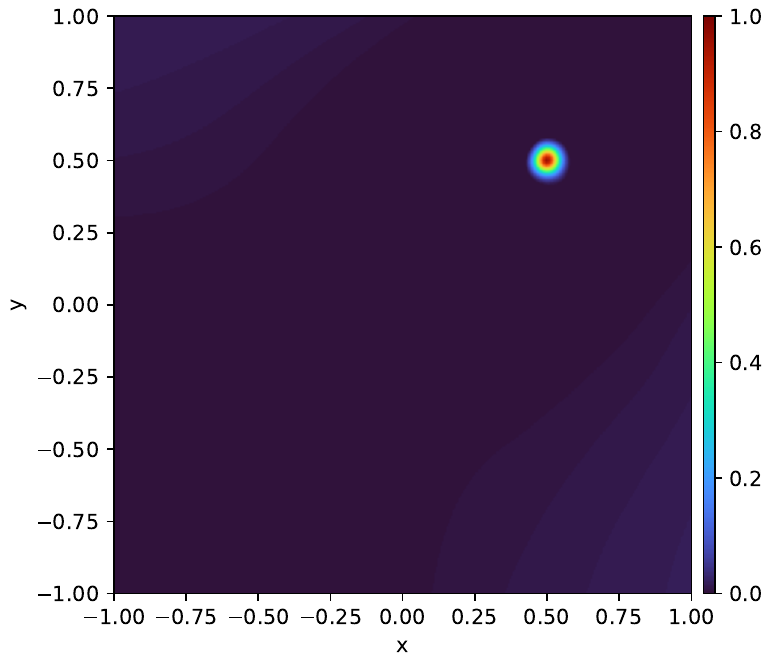}
}
{
    \includegraphics[width=0.2\textwidth]{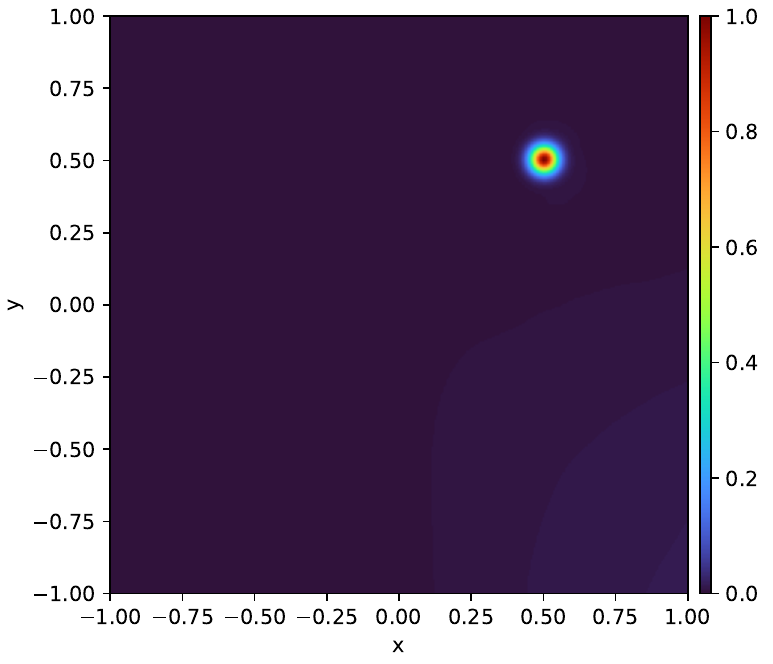}
}
    \subfigure[UNIFORM]
{
    \includegraphics[width=0.2\textwidth]{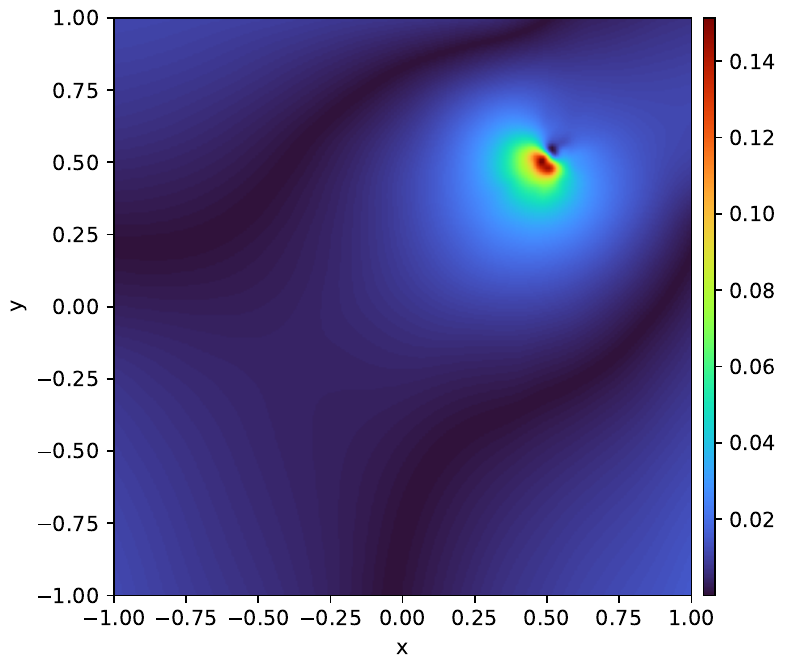}
}
    \subfigure[RAR]
{
    \includegraphics[width=0.2\textwidth]{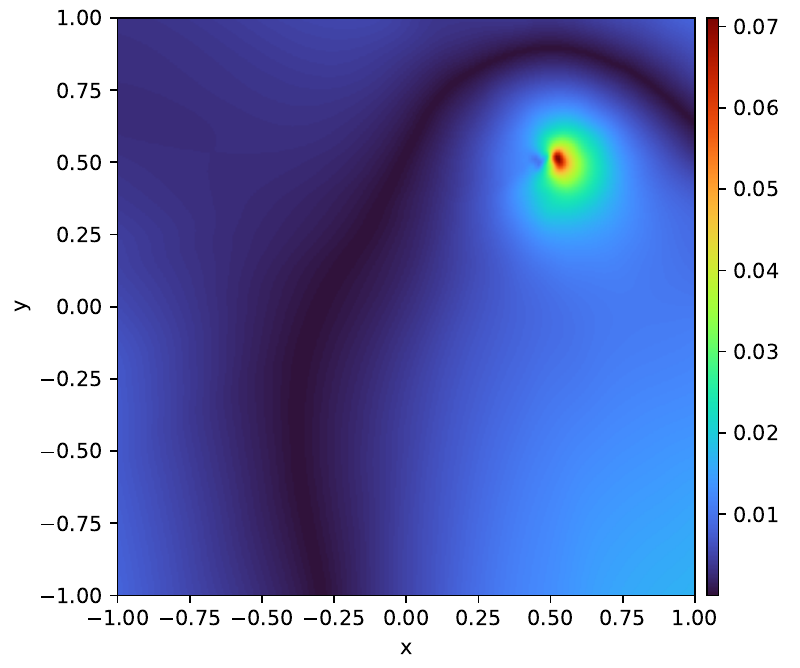}
}
    \subfigure[RAD]
{
    \includegraphics[width=0.2\textwidth]{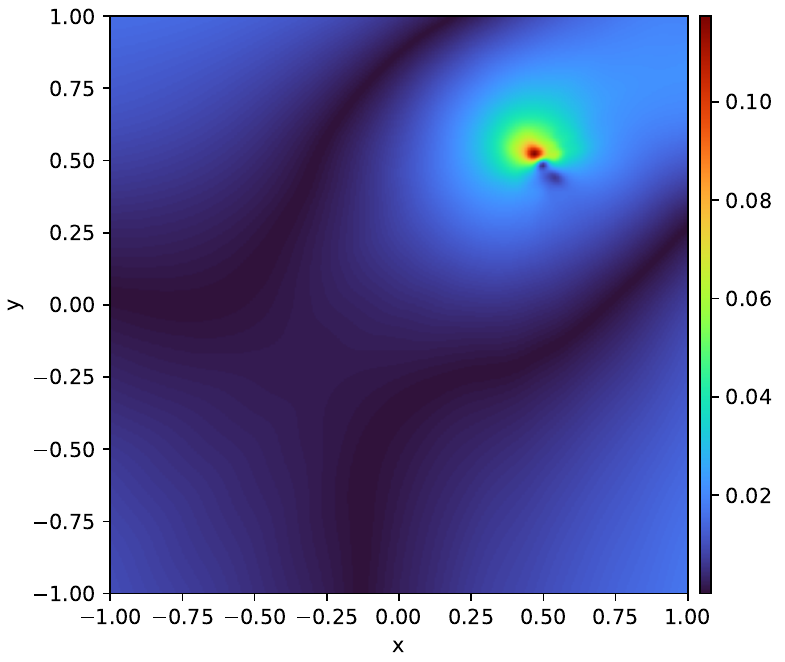}
}
    \subfigure[RL-PINNs]
{
    \includegraphics[width=0.2\textwidth]{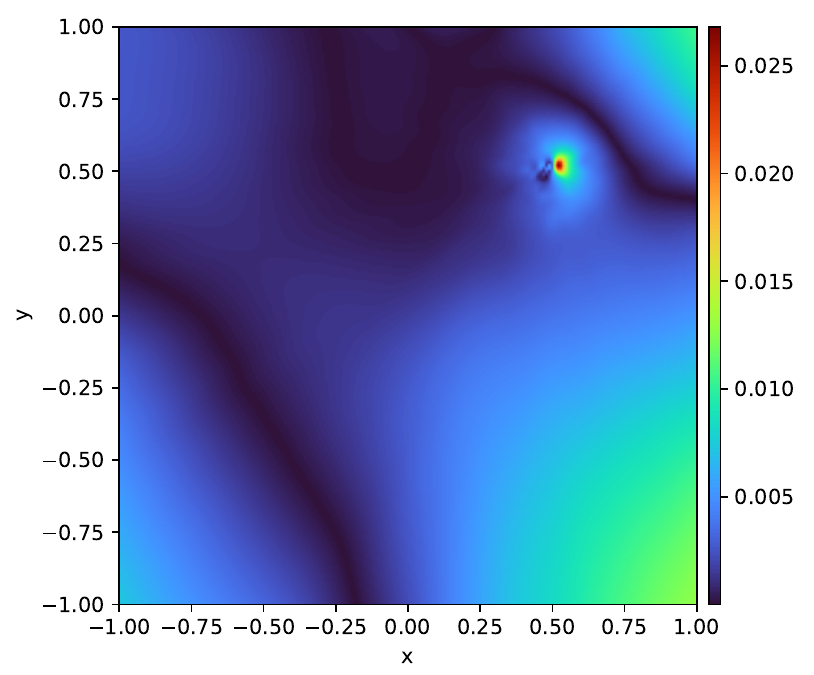}
}

    \caption{\label{fig:peak_error} Single-Peak: Above: The predicted solution. Below: The absolute error of the solution.}
\end{figure}

\begin{figure}[h]
    \centering
    \subfigure[UNIFORM]
{
    \includegraphics[width=0.2\textwidth]{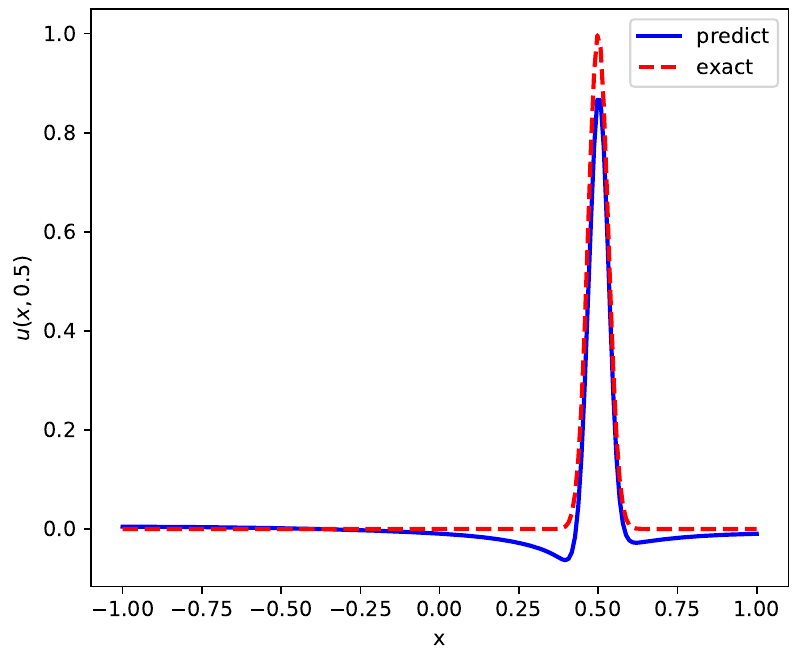}
}
    \subfigure[RAR]
{
    \includegraphics[width=0.2\textwidth]{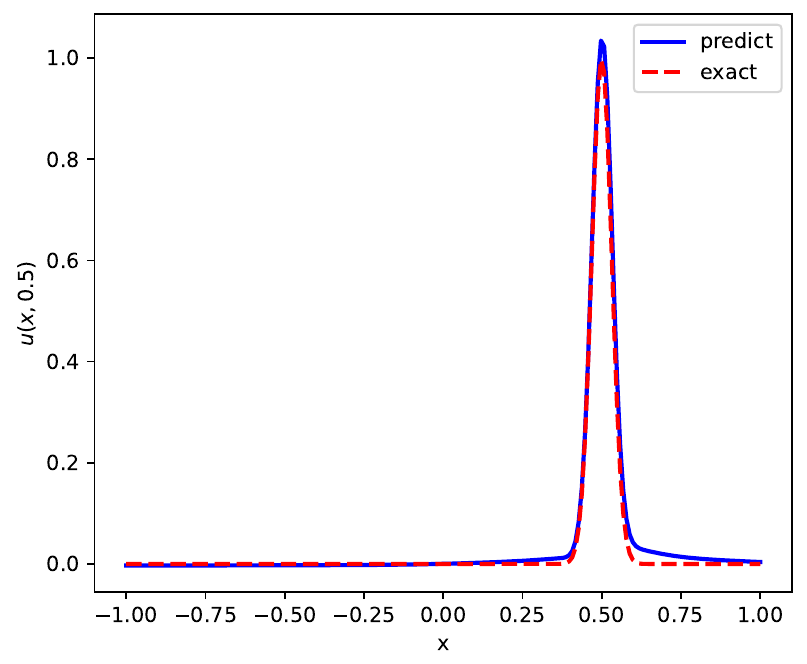}
}
    \subfigure[RAD]
{
    \includegraphics[width=0.2\textwidth]{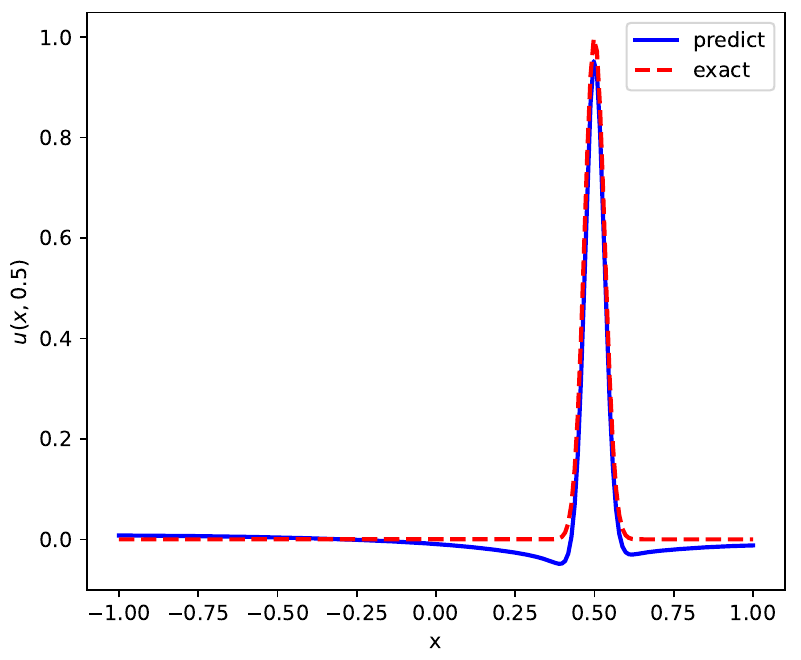}
}
    \subfigure[RL-PINNs]
{
    \includegraphics[width=0.2\textwidth]{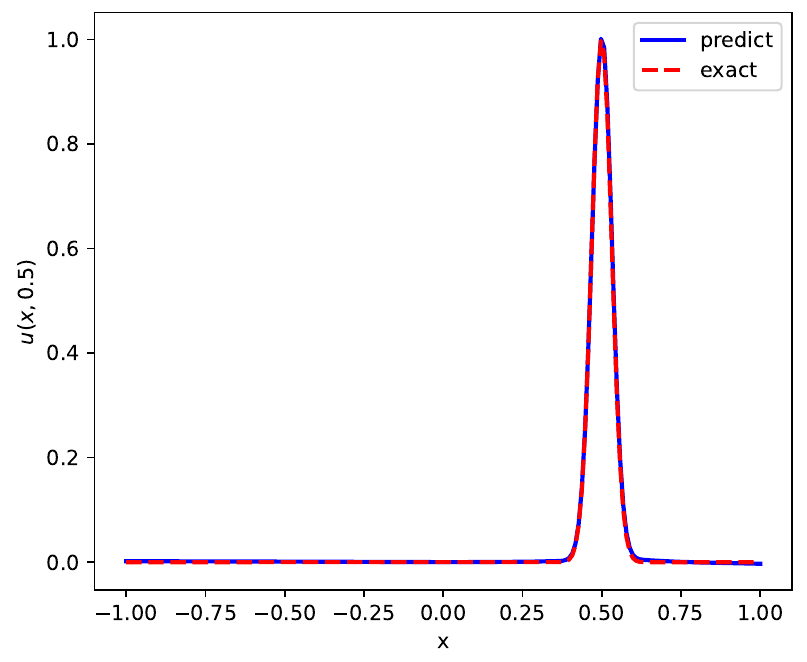}
}
    \caption{\label{fig:peak_half} Single-Peak: The predicted solution at $(x,0.5)$.}
\end{figure}

\begin{equation}\label{eq:11}
    \begin{aligned}
\left\{\begin{array}{l}
-\Delta u(x, y)  =f(x, y), \quad  \text { in } \Omega = [-1,1]^2, \\
u(x, y)  =g(x, y), \quad \text { on } \partial \Omega,
\end{array}\right.
    \end{aligned}
\end{equation}

where the exact solution is a sharply peaked Gaussian function:

\begin{equation}\label{eq:12}
    \begin{aligned}
u(x, y)=\exp^{-500\left[(x-0.5)^2+(y-0.5)^2\right]}.
    \end{aligned}
\end{equation}

The solution exhibits a steep gradient near $(0.5,0.5)$. Implementation details are as follows:

\begin{itemize}
	\item RL-PINNs perform one round of adaptive sampling (retaining 667 high-variation points after thresholding).
	\item Baseline methods (UNIFORM, RAR, RAD) execute five rounds, each adding 200 collocation points (1000 points total).
    \item Post-sampling training: RL-PINNs undergo 25000 iterations. Baselines train for 5000 iterations per round.
\end{itemize}

As shown in Tab.\ref{tab_onepeak}, RL-PINNs achieve a relative $L_2$ error of 0.1462, outperforming UNIFORM (0.4242), RAR (0.2871), and RAD (0.4045) by 65.5\%, 49.1\%, and 63.8\%, respectively. This improvement stems from RL-PINNs’ ability to concentrate samples near the peak, whereas baseline methods distribute points suboptimally, leading to redundant sampling in smooth regions (Fig.\ref{fig:peak_points}).

The total computational time for RL-PINNs is 591.68 seconds. Although its sampling phase is longer (3.32 s compared to 0.14 s for RAR and RAD), the sampling overhead accounts for only 0.56\% of the total runtime. This indicates that the sampling cost of RL-PINNs is negligible relative to the overall training time of the PINN. As illustrated in Fig.\ref{fig:peak_error}, RL-PINNs also achieve lower absolute errors across the domain, particularly near the peak regions.

\begin{table}[h]
\caption{Results of Single-Peak case}
\label{tab_onepeak}
\centering
\begin{tabular}{cccc}  
\toprule
Sigle-Peak &  Total sampling time & Total PINNs training time & $L_2$ \\  
\midrule
UNIFORM & 6.35e-4 s & 611.02 s & 0.4242 \\  
RAR & 0.14 s & 613.70 s & 0.2871 \\  
RAD & 0.14 s & 616.12 s & 0.4045 \\  
RL-PINNs & 3.32 s & 588.36 s & 0.1462 \\  
\bottomrule
\end{tabular}
\end{table}

\subsubsection{Dual-Peak Case}
\ 
\newline

We extend the benchmark to a Dual-Peak solution:

\begin{equation}\label{eq:13}
    \begin{aligned}
u(x, y)=\exp^{-500\left[(x+0.5)^2+(y+0.5)^2\right]-500\left[(x-0.5)^2+(y-0.5)^2\right]},
    \end{aligned}
\end{equation}

\begin{figure}[h]
    \centering
{
    \includegraphics[width=0.4\textwidth]{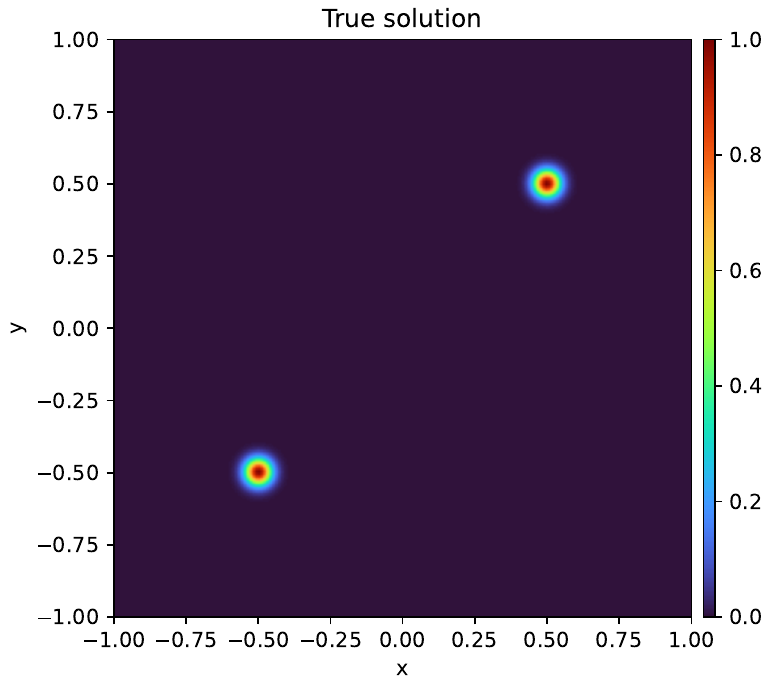}
}
    \caption{\label{fig:peak2_true_u} Dual-Peak: The exact solution.}
\end{figure}

with peaks  at $(-0.5,-0.5)$ and $(0.5,0.5)$. Implementation details are as follows:

\begin{itemize}
	\item RL-PINNs perform one round of adaptive sampling (retaining 1271 high-variation points).
    \item Baseline execute five rounds, each adding 400 collocation points (2000 points total).	
    \item Training protocols match the Single-Peak case.
\end{itemize}

\begin{figure}[h]
    \centering
    \subfigure[UNIFORM]
{
    \includegraphics[width=0.35\textwidth]{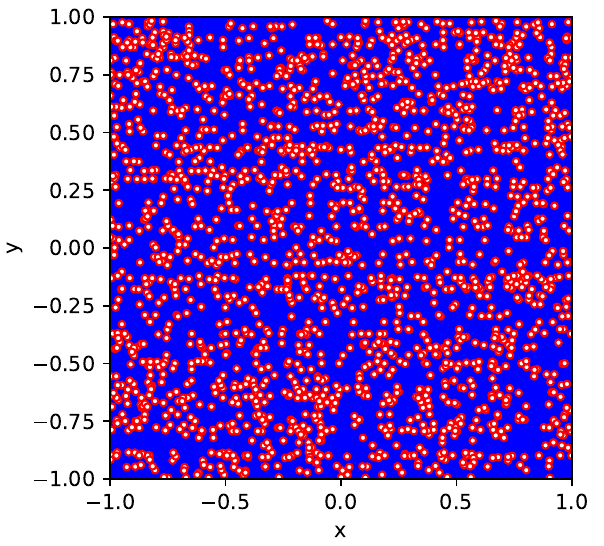}
}
    \subfigure[RAR]
{
    \includegraphics[width=0.35\textwidth]{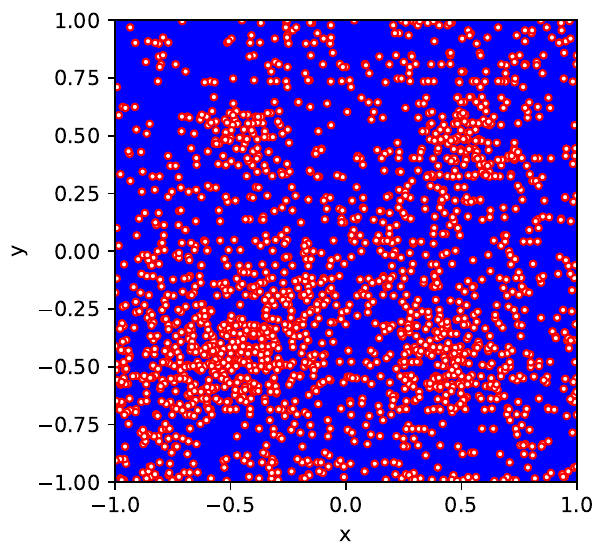}
}
    \subfigure[RAD]
{
    \includegraphics[width=0.35\textwidth]{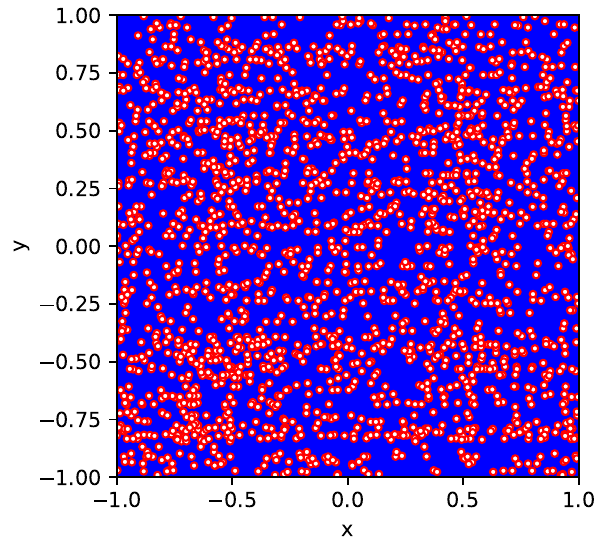}
}
    \subfigure[RL-PINNs]
{
    \includegraphics[width=0.35\textwidth]{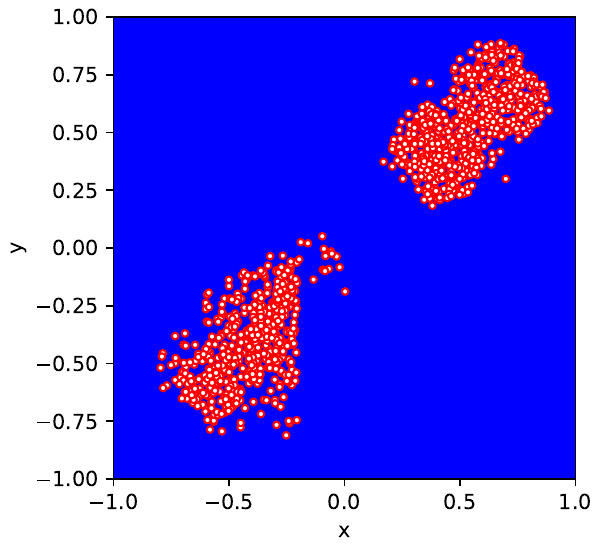}
}
    \caption{\label{fig:peak2_points} Dual-Peak: Cumulative sampling of UNIFORM, RAR, RAD and one-time sampling of RL-PINNs.}
\end{figure}

RL-PINNs achieve a relative $L_2$ error of 0.1878 (Tab.\ref{tab_twopeak}), reducing errors by 48.7\%, 82.8\%, and 78.2\% compared to RAR (0.3659), RAD (1.0889), and UNIFORM (0.8624), respectively. Fig.\ref{fig:peak2_points} demonstrates that RL-PINNs dynamically allocate points near both peaks, whereas sampling points of RAR and RAD are too scattered.

\begin{figure}[h]
    \centering
{
    \includegraphics[width=0.2\textwidth]{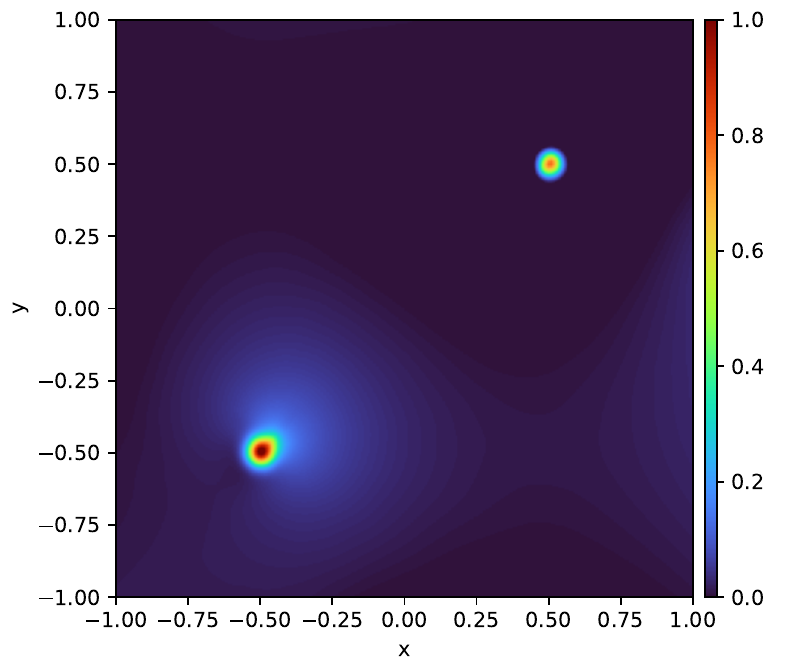}
}
{
    \includegraphics[width=0.2\textwidth]{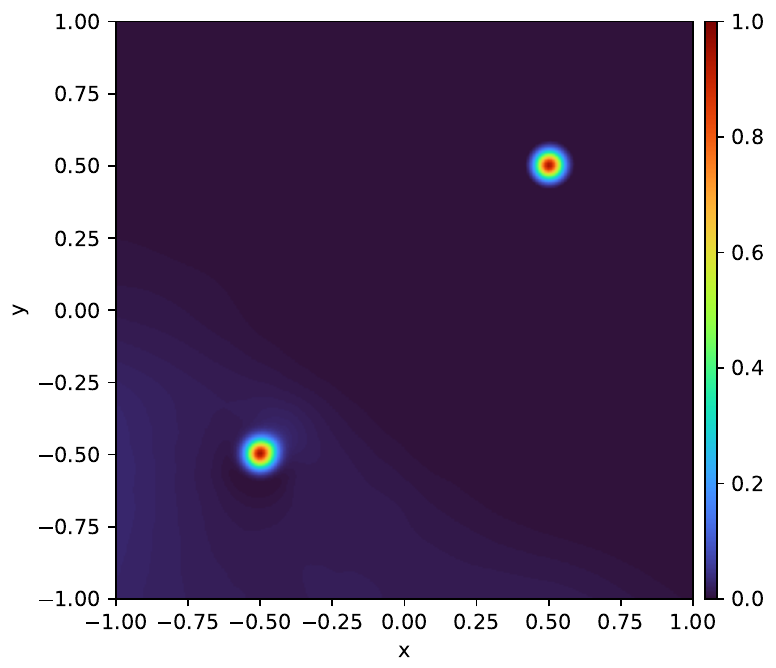}
}
{
    \includegraphics[width=0.2\textwidth]{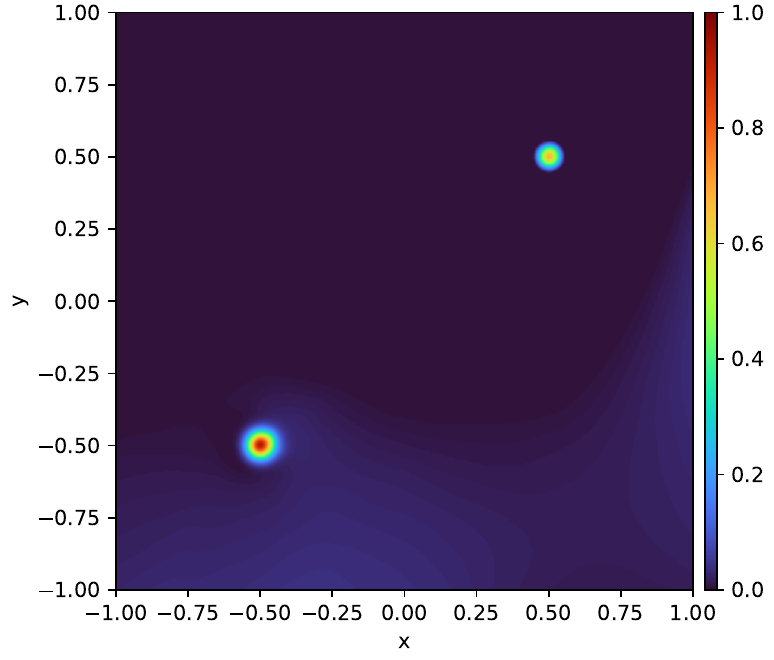}
}
{
    \includegraphics[width=0.2\textwidth]{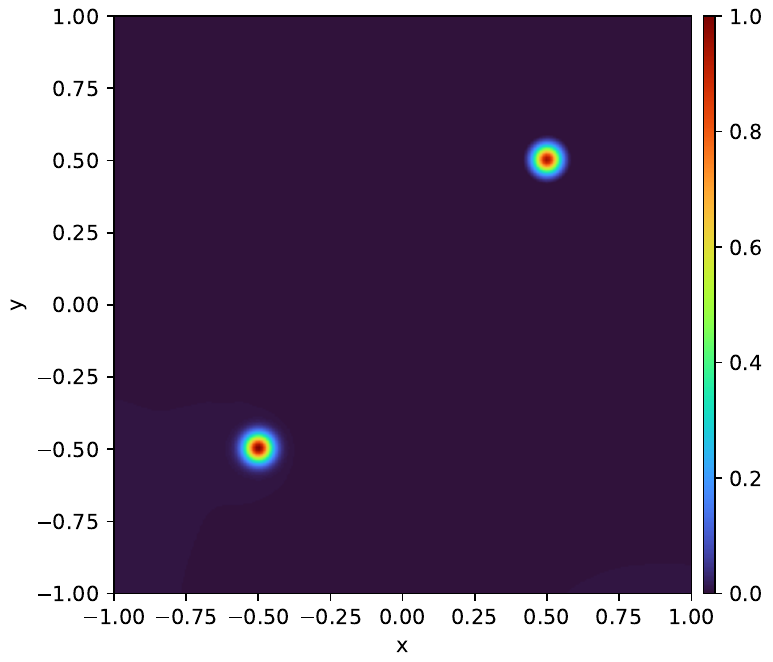}
}
    \subfigure[UNIFORM]
{
    \includegraphics[width=0.2\textwidth]{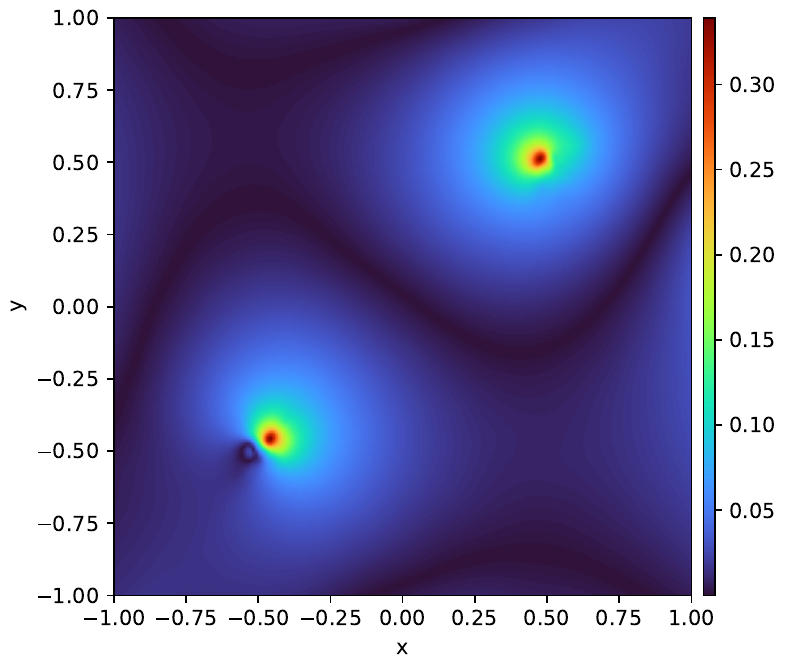}
}
    \subfigure[RAR]
{
    \includegraphics[width=0.2\textwidth]{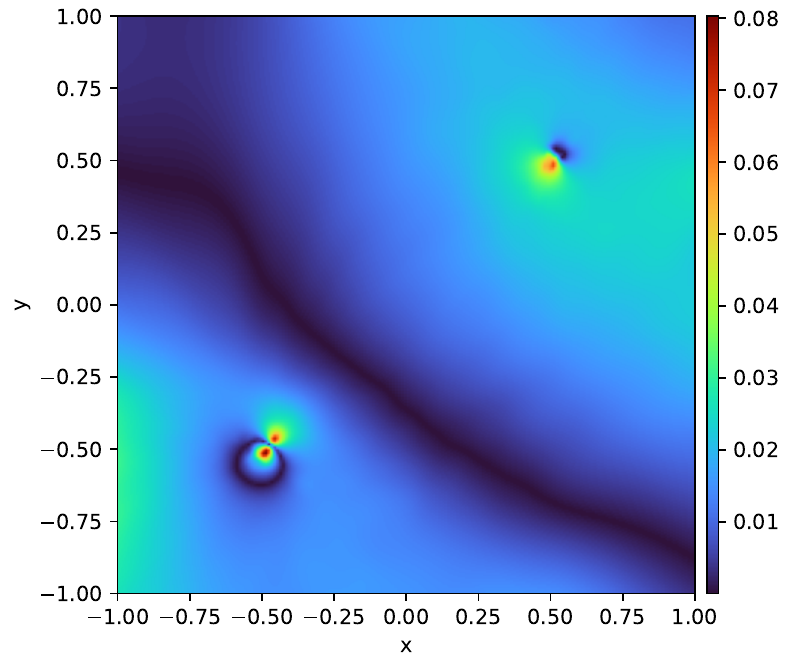}
}
    \subfigure[RAD]
{
    \includegraphics[width=0.2\textwidth]{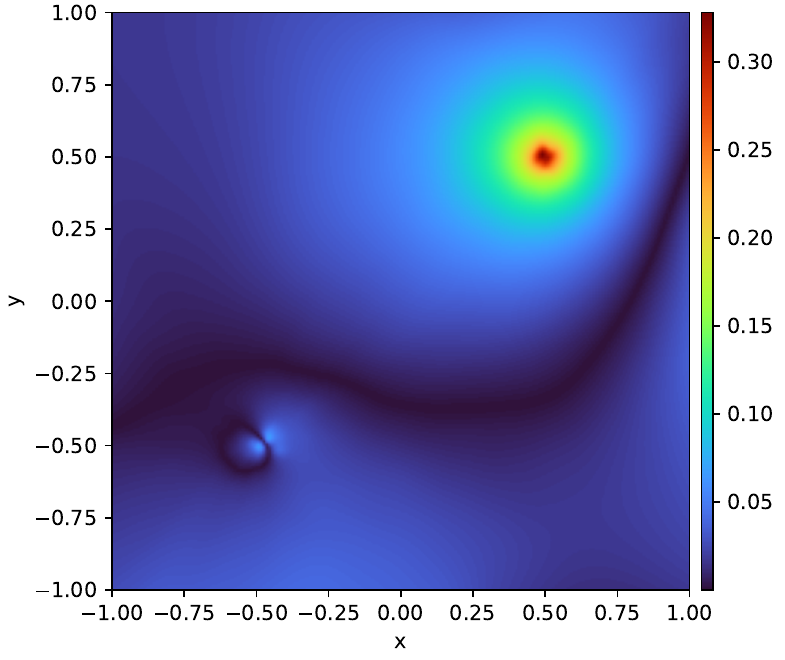}
}
    \subfigure[RL-PINNs]
{
    \includegraphics[width=0.2\textwidth]{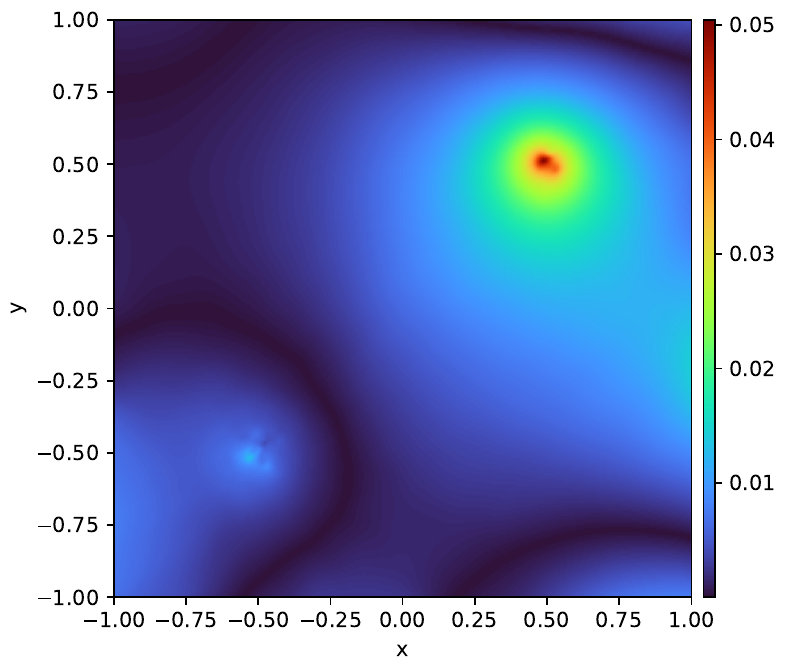}
}
    \caption{\label{fig:peak2_error} Dual-Peak: Above: The predicted solution. Below: The absolute error of the solution.}
\end{figure}

\begin{figure}[h]
    \centering
{
    \includegraphics[width=0.2\textwidth]{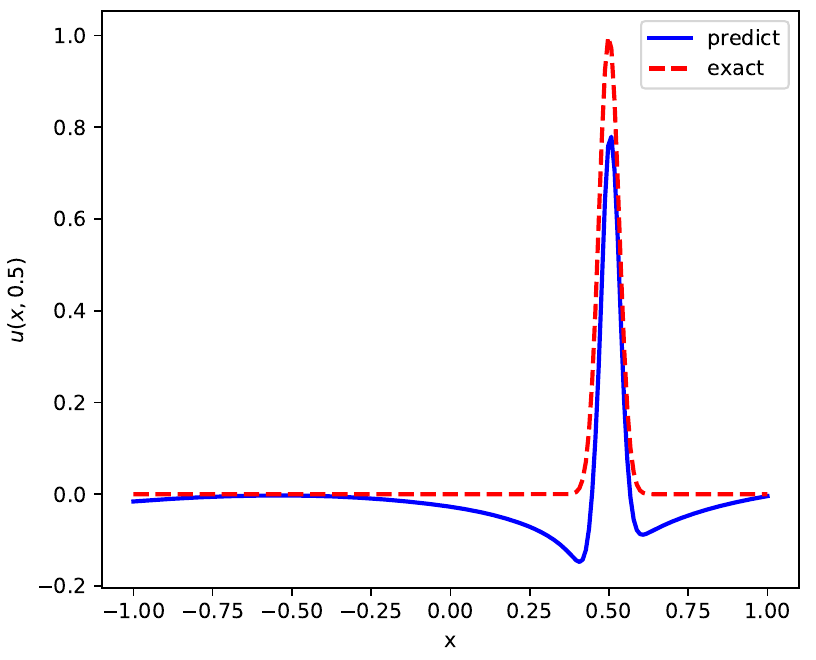}
}
{
    \includegraphics[width=0.2\textwidth]{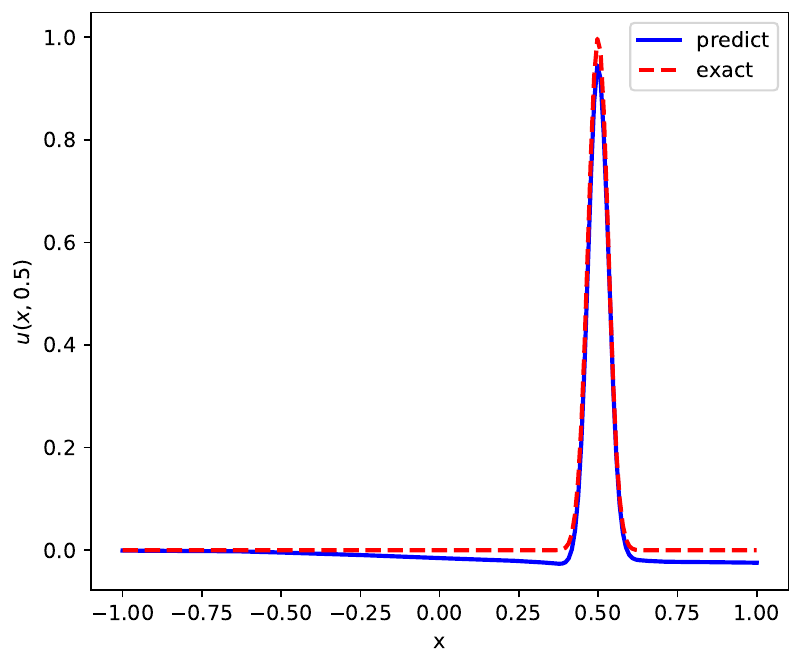}
}
{
    \includegraphics[width=0.2\textwidth]{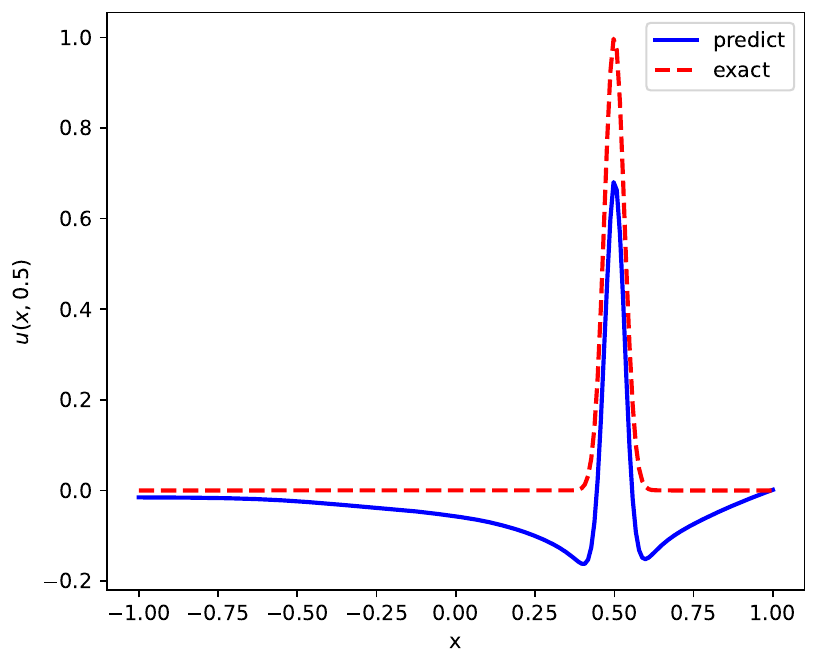}
}
{
    \includegraphics[width=0.2\textwidth]{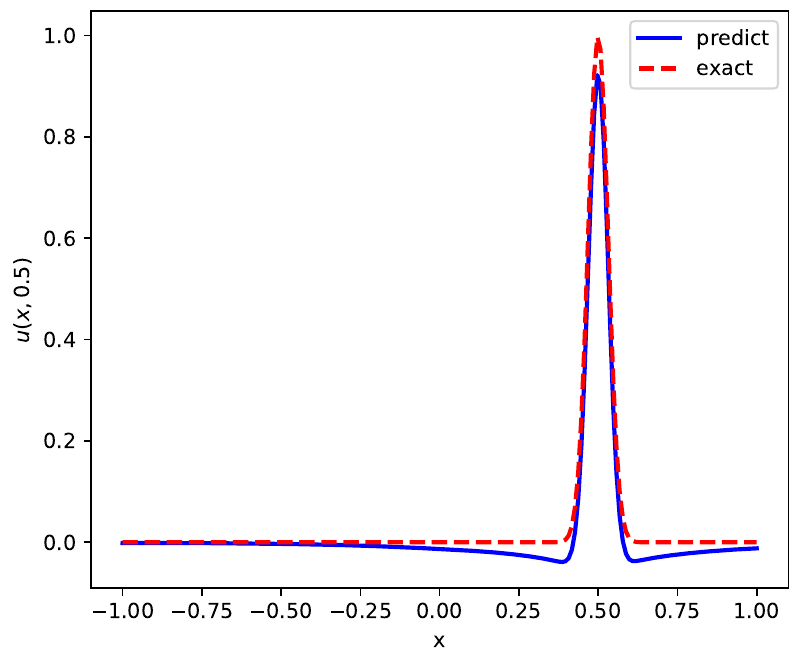}
}
    \subfigure[UNIFORM]
{
    \includegraphics[width=0.2\textwidth]{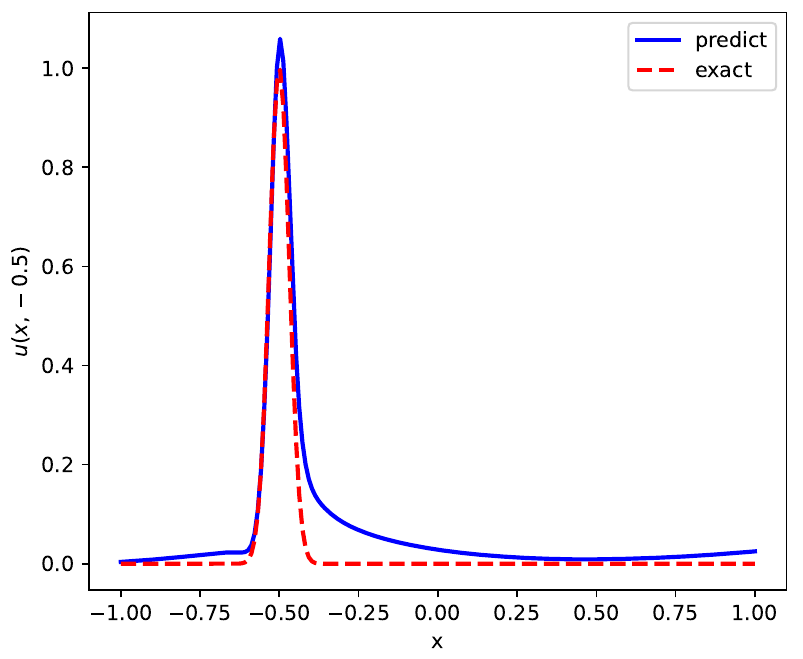}
}
    \subfigure[RAR]
{
    \includegraphics[width=0.2\textwidth]{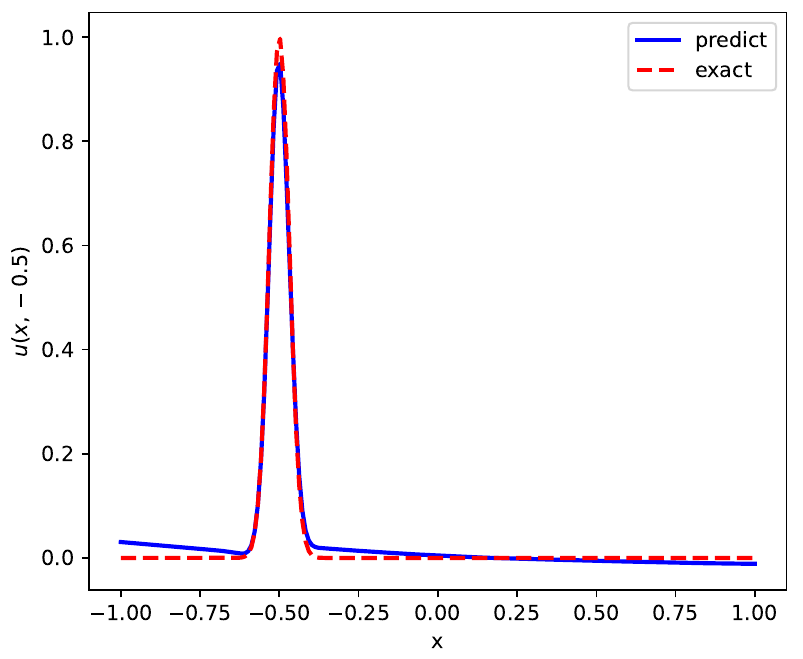}
}
    \subfigure[RAD]
{
    \includegraphics[width=0.2\textwidth]{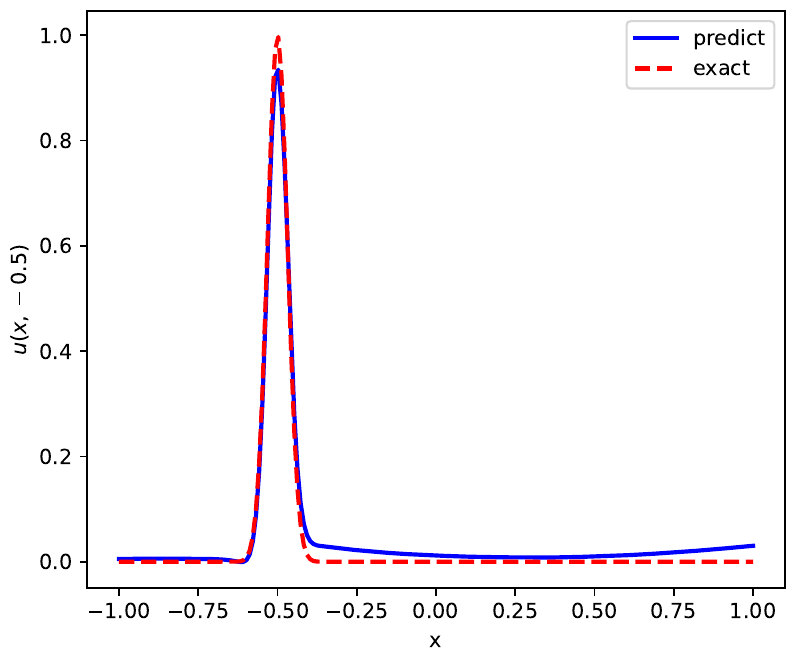}
}
    \subfigure[RL-PINNs]
{
    \includegraphics[width=0.2\textwidth]{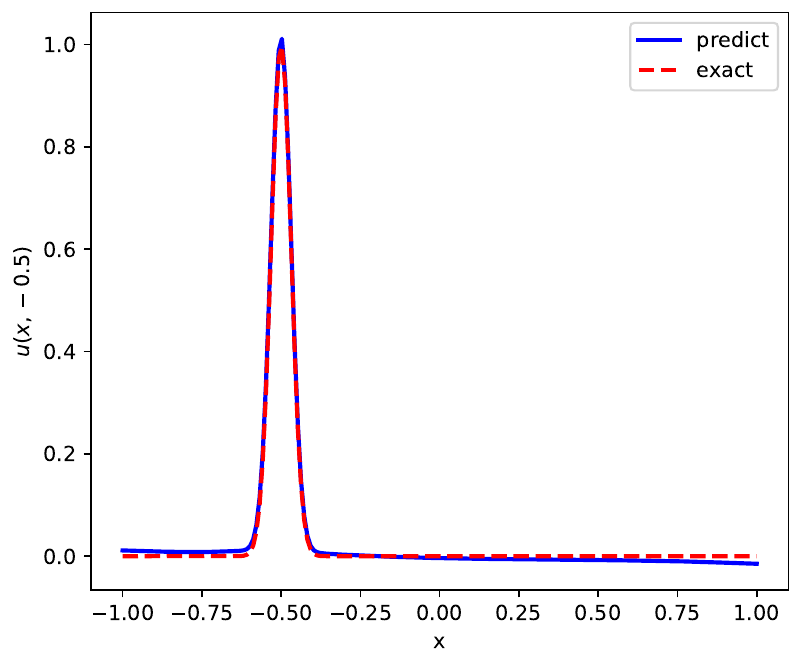}
}
    \caption{\label{fig:peak2_half} Dual-Peak: The predicted solution at $(x,0.5)$ and $(x,-0.5)$.}
\end{figure}

\begin{table}[h]
\caption{Results of Dual-Peak case}
\label{tab_twopeak}
\centering
\begin{tabular}{cccc}  
\toprule
Dual-Peak &  Total sampling time & Total PINNs training time & $L_2$ \\  
\midrule
UNIFORM & 6.59e-3 s & 652.76 s & 0.8624 \\  
RAR & 0.26 s & 663.94 s & 0.3659 \\  
RAD & 0.27 s & 650.68 s & 1.0889 \\  
RL-PINNs & 8.68 s & 614.29 s & 0.1878 \\  
\bottomrule
\end{tabular}
\end{table}

\subsection{Burgers' equation.}
\ 
\newline

We evaluate RL-PINNs on the one-dimensional viscous Burgers' equation, a canonical nonlinear PDE with sharp gradient evolution.  The problem is formulated as:

\begin{figure}[h]
    \centering
{
    \includegraphics[width=0.5\textwidth]{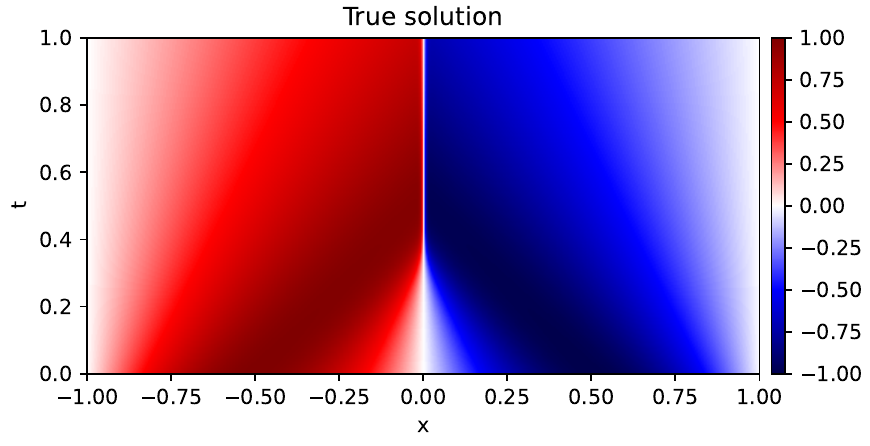}
}
    \caption{\label{fig:burgers_true_u} Burgers': The exact solution.}
\end{figure}

\begin{figure}[h]
    \centering
    \subfigure[UNIFORM]
{
    \includegraphics[width=0.35\textwidth]{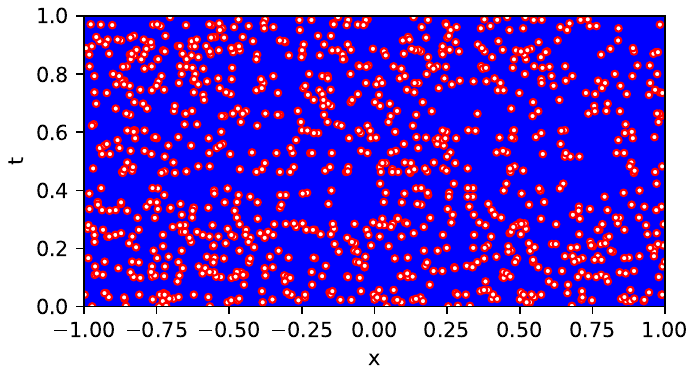}
}
    \subfigure[RAR]
{
    \includegraphics[width=0.35\textwidth]{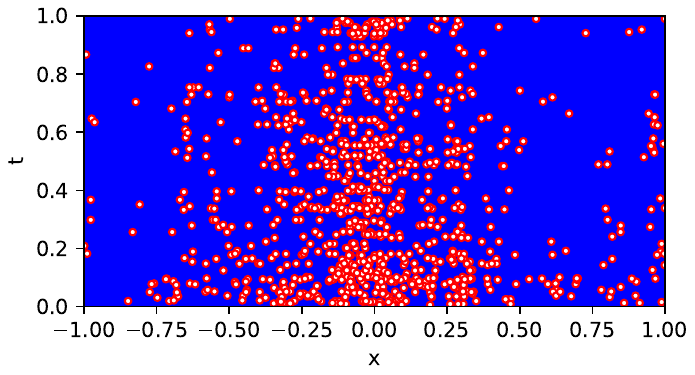}
}
    \subfigure[RAD]
{
    \includegraphics[width=0.35\textwidth]{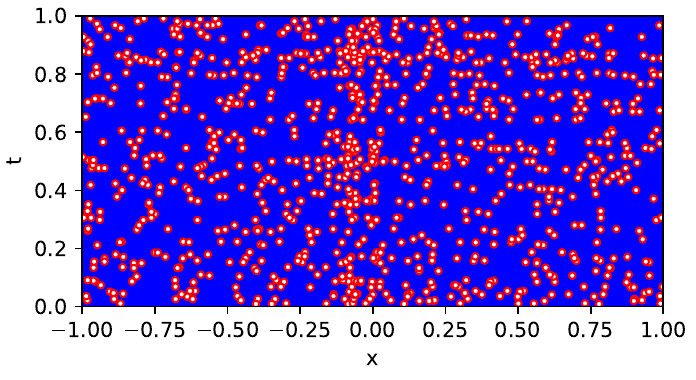}
}
    \subfigure[RL-PINNs]
{
    \includegraphics[width=0.35\textwidth]{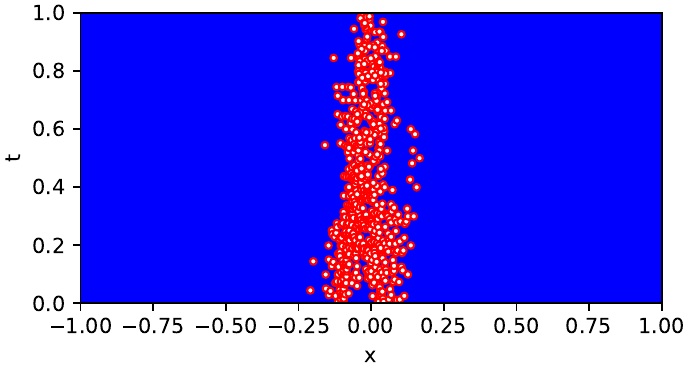}
}
    \caption{\label{fig:burgers_points} Burgers': Cumulative sampling of UNIFORM, RAR, RAD and one-time sampling of RL-PINNs.}
\end{figure}

\begin{figure}[h]
    \centering
    \subfigure[UNIFORM]
{
    \includegraphics[width=0.35\textwidth]{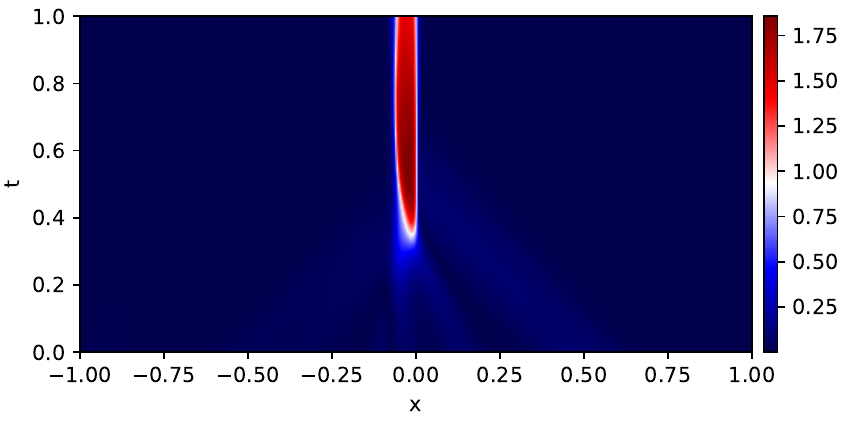}
}
    \subfigure[RAR]
{
    \includegraphics[width=0.35\textwidth]{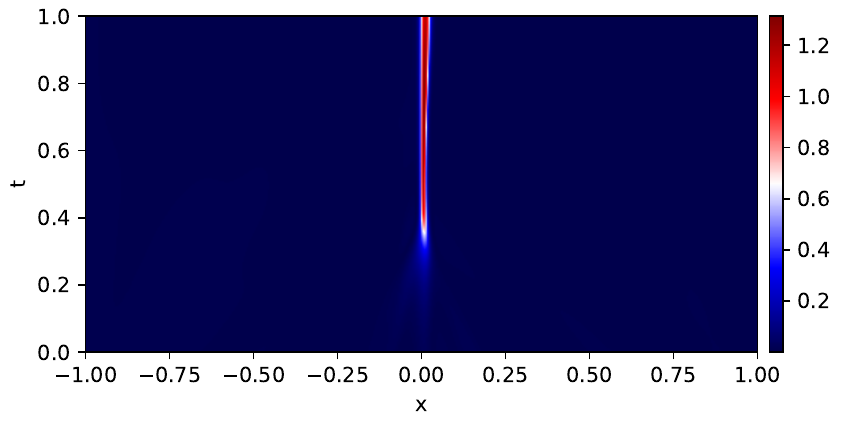}
}
    \subfigure[RAD]
{
    \includegraphics[width=0.35\textwidth]{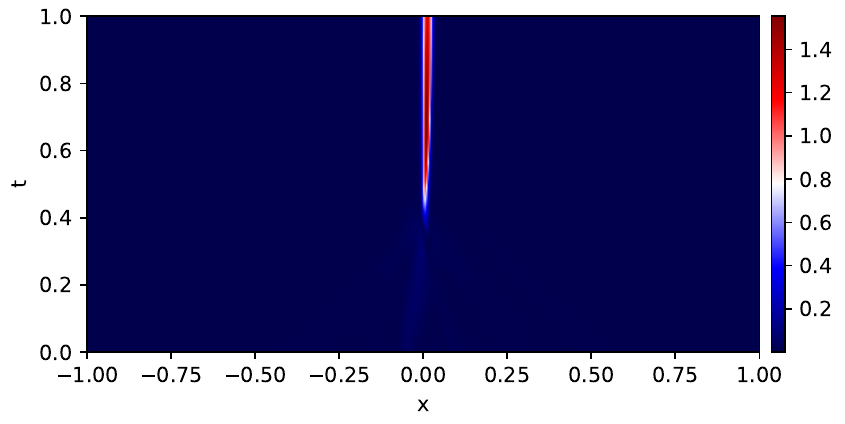}
}
    \subfigure[RL-PINNs]
{
    \includegraphics[width=0.35\textwidth]{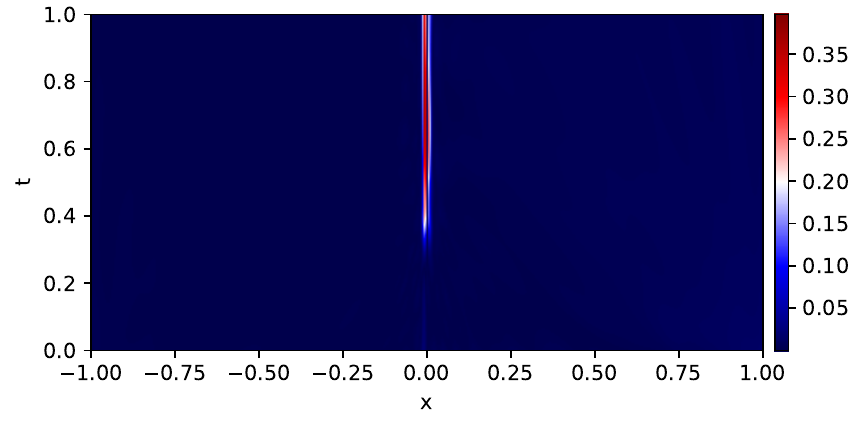}
}
    \caption{\label{fig:burgers_error} Burgers': The absolute error of the solution.}
\end{figure}

\begin{figure}[h]
    \centering
    \subfigure[UNIFORM]
{
    \includegraphics[width=0.2\textwidth]{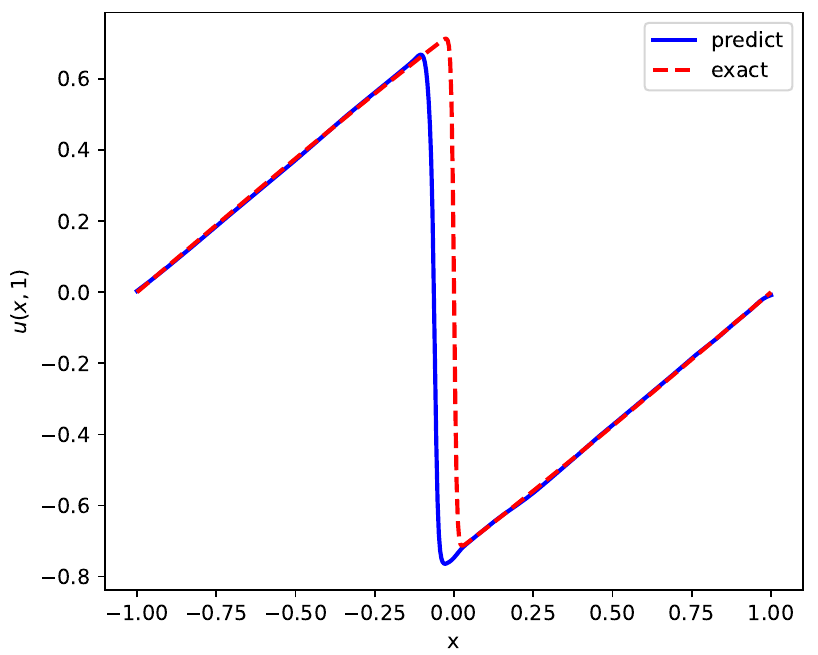}
}
    \subfigure[RAR]
{
    \includegraphics[width=0.2\textwidth]{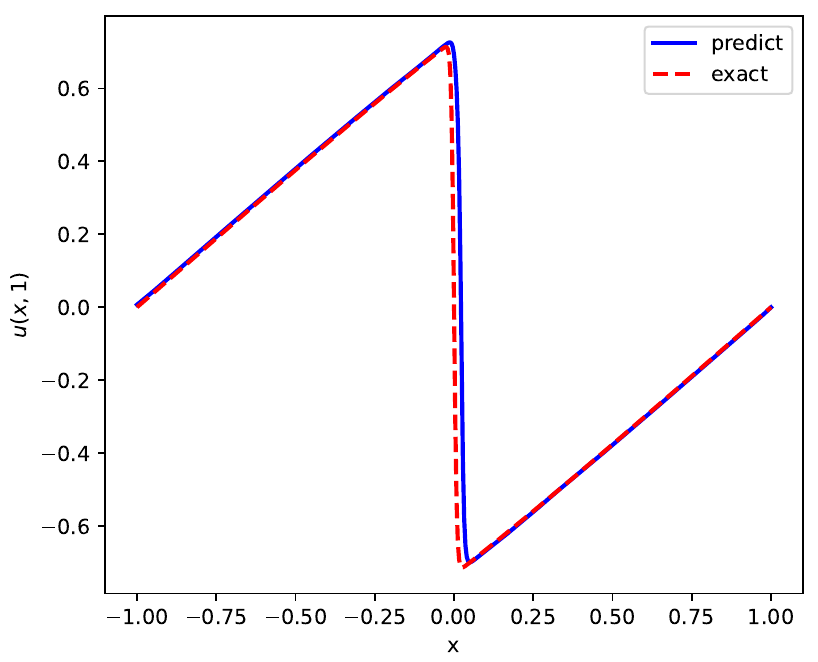}
}
    \subfigure[RAD]
{
    \includegraphics[width=0.2\textwidth]{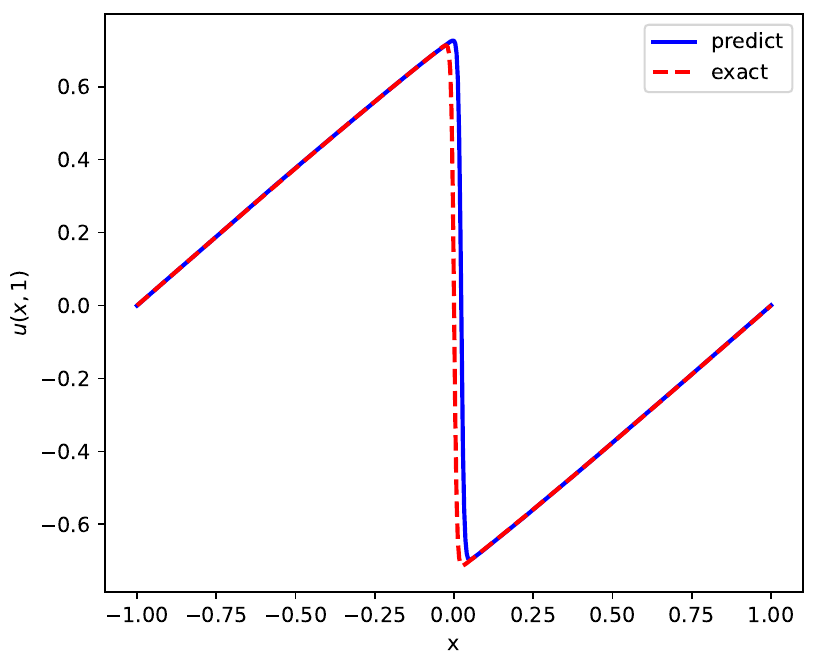}
}
    \subfigure[RL-PINNs]
{
    \includegraphics[width=0.2\textwidth]{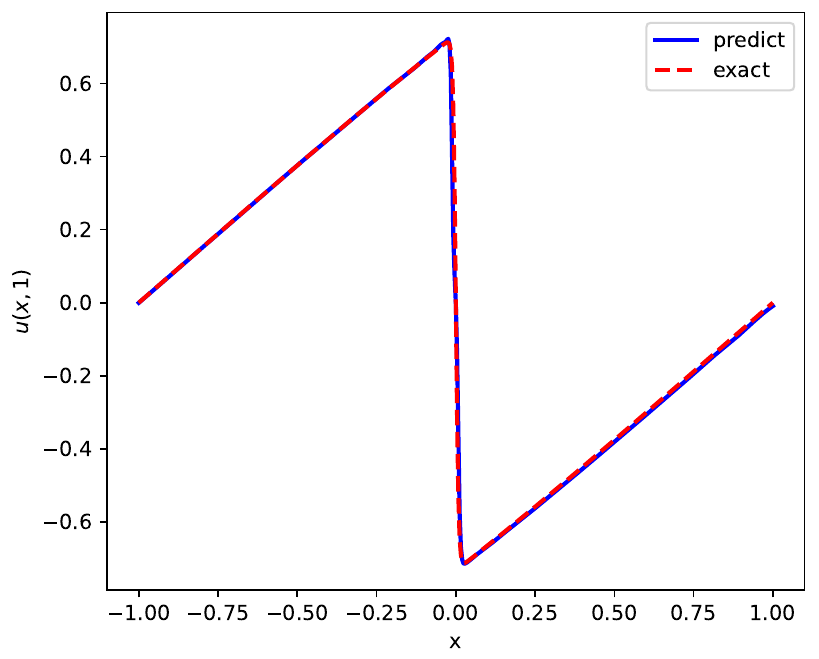}
}
    \caption{\label{fig:burgers_error_t1}  Burgers': The predicted solution $t=1$.}
\end{figure}

\begin{equation}\label{eq:14}
    \begin{aligned}
\left\{\begin{array}{l}
u_t+u u_x-\frac{0.01}{\pi} u_{x x}=0, \\
u(x, 0)=-\sin (\pi x), \\
u(t,-1)=u(t, 1)=0,
\end{array}\right.
    \end{aligned}
\end{equation}

where $(x,t) \in [-1,1] \times [0,1]$. The solution develops a steep gradient near $x=0$ as $t\rightarrow 1$, posing significant challenges for traditional adaptive sampling methods due to the need for localized resolution of the shock front. Implementation details are as follows:

\begin{itemize}
	\item RL-PINNs perform one round of adaptive sampling (retaining 634 high-variation points).
	\item Baseline methods (UNIFORM, RAR, RAD) execute five rounds, each adding 200 collocation points (1000 points total).
    \item Post-sampling training employs 25000 Adam iterations followed by 25000 L-BFGS fine-tuning steps for RL-PINNs, compared to 5000 Adam + 5000 L-BFGS steps per round for baselines.
\end{itemize}

Tab.\ref{tab_burgers} summarizes the computational performance. RL-PINNs achieve a relative $L_2$ error of 0.0534, outperforming UNIFORM (0.2896), RAR (0.1323), and RAD (0.1474) by 81.5\%, 59.6\%, and 63.8\%, respectively. 

Fig.\ref{fig:burgers_points} illustrates collocation point distributions. RL-PINNs concentrate samples near $x=0$, where solution gradients sharpen temporally, while baselines exhibit uniform distributions inadequate for shock front resolution.

Absolute error comparisons (Fig.\ref{fig:burgers_error}) demonstrate RL-PINNs' superior accuracy near $x=0$,  Fig.\ref{fig:burgers_error_t1} further visualizes the solution profile at $t=1$, demonstrating RL-PINNs’ precise alignment with the exact solution.

\begin{table}[h]
\caption{Results of Burgers' equation}
\label{tab_burgers}
\centering
\begin{tabular}{cccc}  
\toprule
Burgers' &  Total sampling time & Total PINNs training time & $L_2$ \\  
\midrule
UNIFORM & 8.61e-4 s & 1322.65 s & 0.2896 \\  
RAR & 0.18 s & 1326.59 s & 0.1323 \\  
RAD & 0.18 s & 1328.56 s & 0.1474 \\  
RL-PINNs & 16.18 s & 1304.39 s & 0.0534 \\  
\bottomrule
\end{tabular}
\end{table}

\subsection{ Time‑Dependent Wave Equation}
\ 
\newline

We further validate RL-PINNs on a one-dimensional time-dependent wave equation featuring multi-modal dynamics and discontinuous propagation. The governing equations are:

\begin{figure}[h]
    \centering
{
    \includegraphics[width=0.4\textwidth]{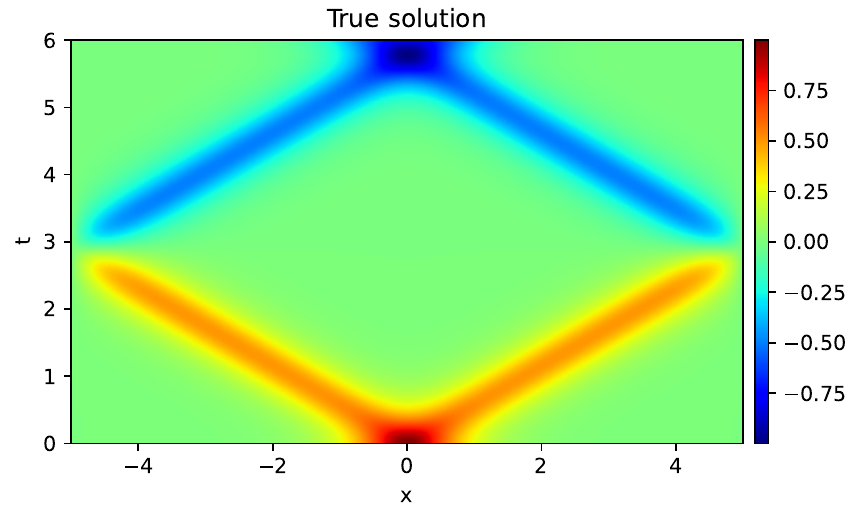}
}
    \caption{\label{fig:wave_true_u} Wave: The exact solution.}
\end{figure}

\begin{figure}[h]
    \centering
    \subfigure[UNIFORM]
{
    \includegraphics[width=0.35\textwidth]{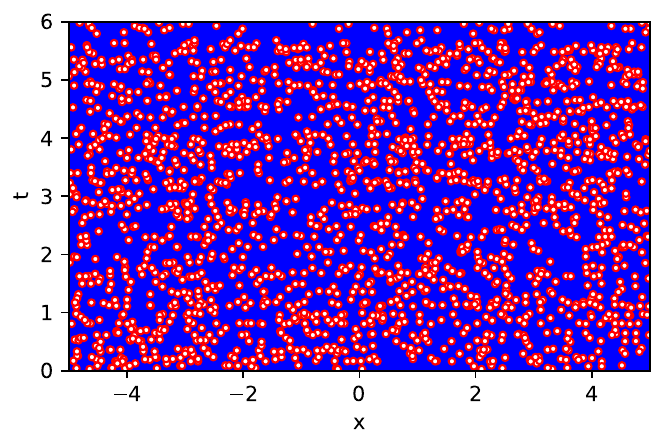}
}
    \subfigure[RAR]
{
    \includegraphics[width=0.35\textwidth]{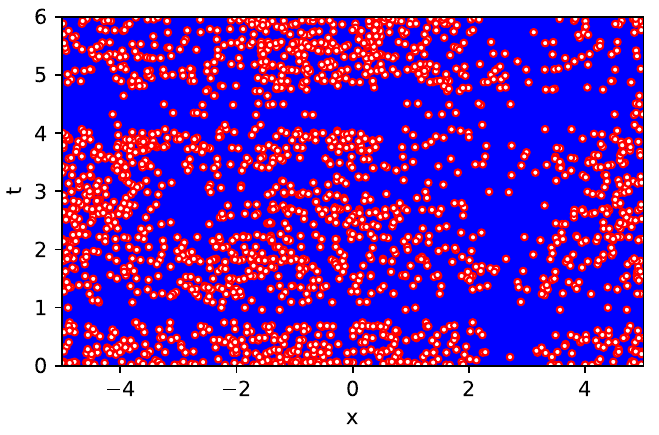}
}
    \subfigure[RAD]
{
    \includegraphics[width=0.35\textwidth]{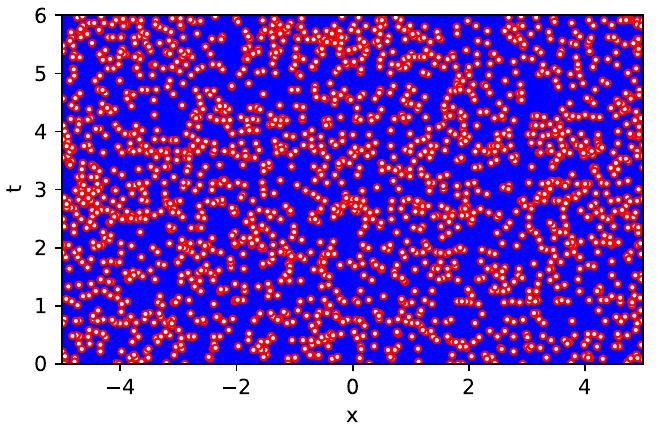}
}
    \subfigure[RL-PINNs]
{
    \includegraphics[width=0.35\textwidth]{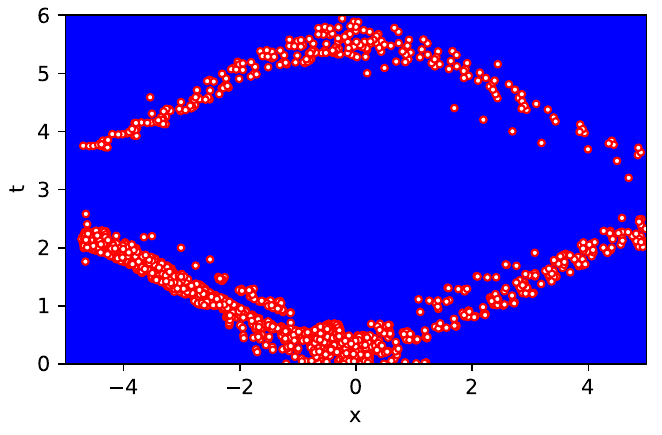}
}
    \caption{\label{fig:wave_points} Wave: Cumulative sampling of UNIFORM, RAR, RAD and one-time sampling of RL-PINNs.}
\end{figure}

\begin{figure}[h]
    \centering

{
    \includegraphics[width=0.2\textwidth]{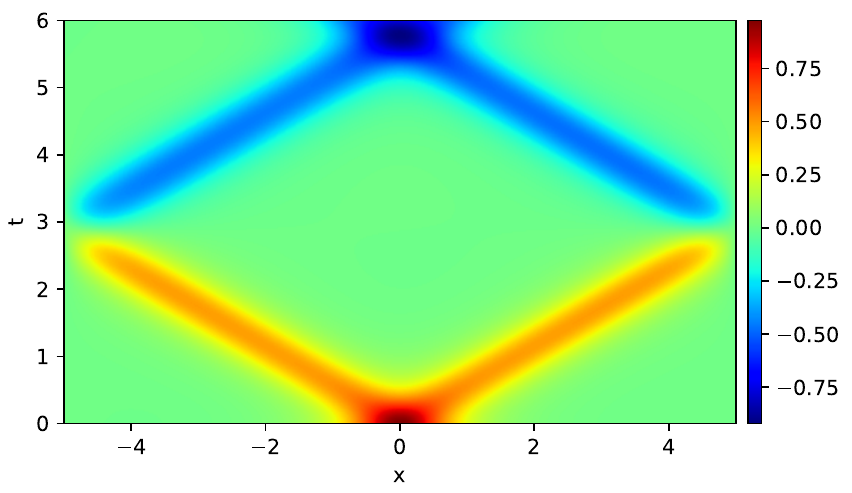}
}
{
    \includegraphics[width=0.2\textwidth]{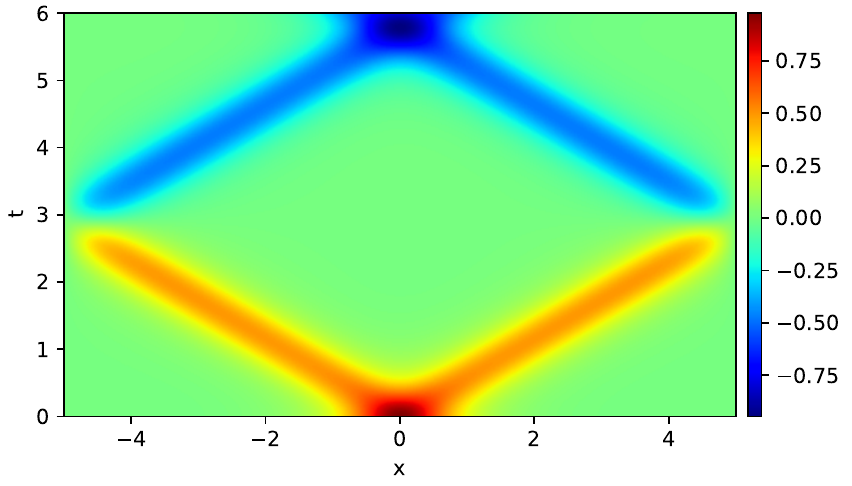}
}
{
    \includegraphics[width=0.2\textwidth]{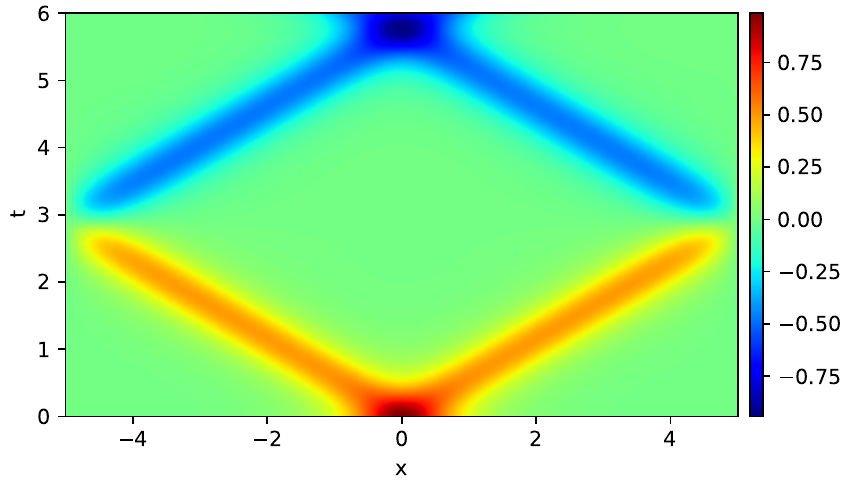}
}
{
    \includegraphics[width=0.2\textwidth]{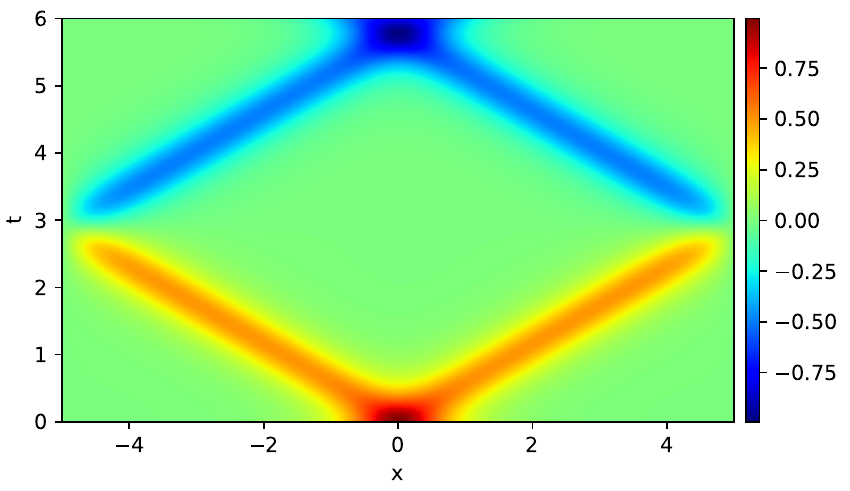}
}
    \subfigure[UNIFORM]
{
    \includegraphics[width=0.2\textwidth]{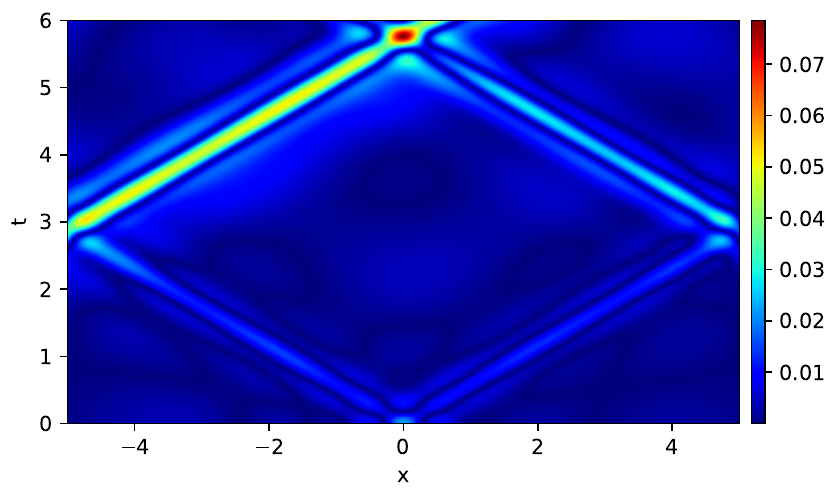}
}
    \subfigure[RAR]
{
    \includegraphics[width=0.2\textwidth]{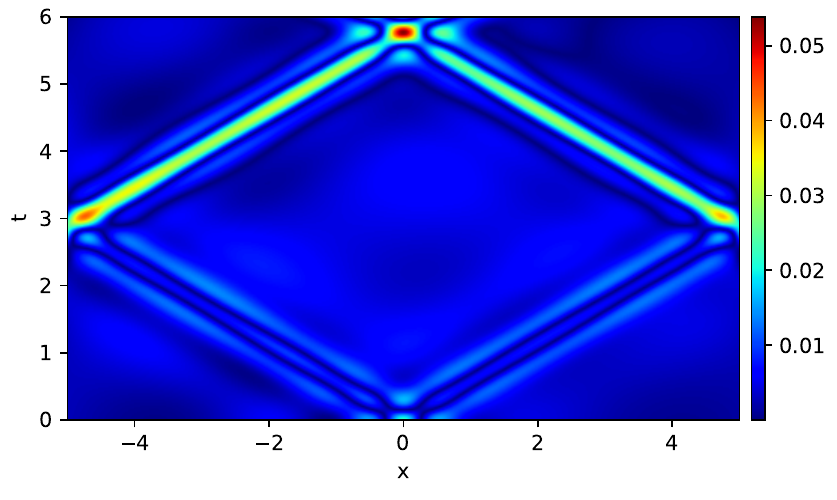}
}
    \subfigure[RAD]
{
    \includegraphics[width=0.2\textwidth]{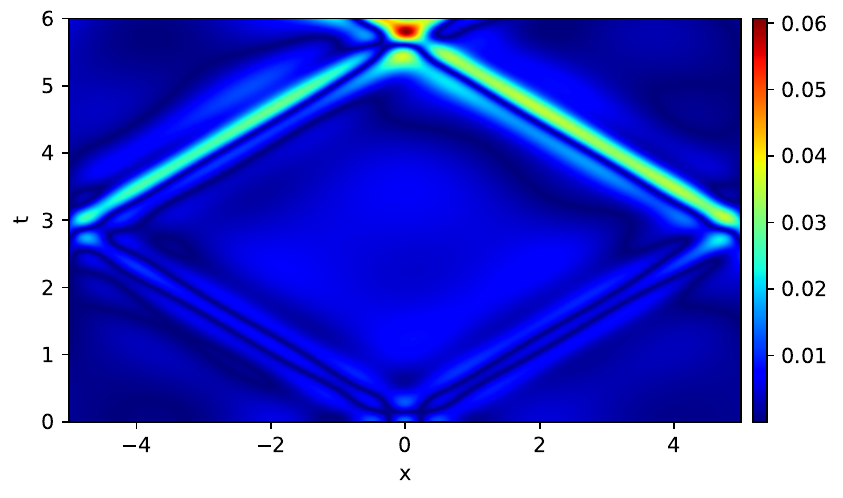}
}
    \subfigure[RL-PINNs]
{
    \includegraphics[width=0.2\textwidth]{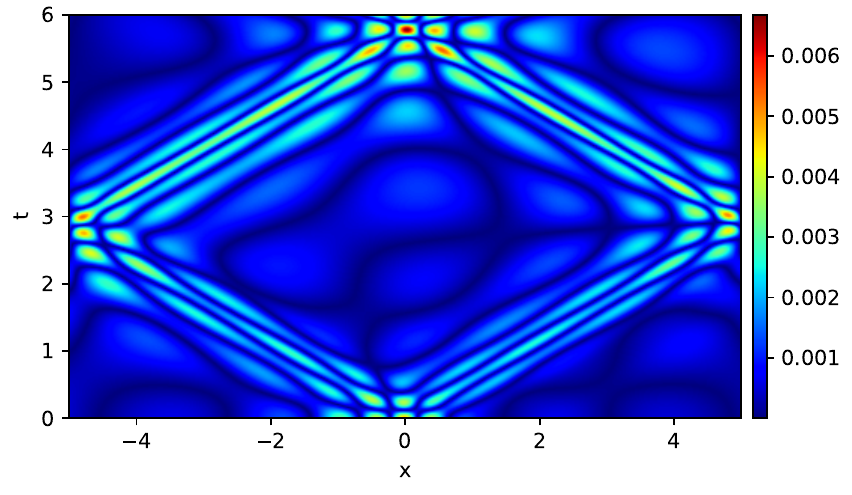}
}
    \caption{\label{fig:wave_error} Wave: Above: The predicted solution. Below: The absolute error of the solution.}
\end{figure}

\begin{equation}\label{eq:15}
    \begin{aligned}
\left\{\begin{array}{l}
\frac{\partial^2 u}{\partial t^2}-3 \frac{\partial^2 u^2}{\partial x}=0,  \\
u(x, 0)=\frac{1}{\cosh (2 x)}-\frac{0.5}{\cosh (2(x-10))}-\frac{0.5}{\cosh (2(x+10))}, \\
\frac{\partial u}{\partial t}(x, 0)=0, \\
u(-5,t)=u(5, t)=0,
\end{array}\right.
    \end{aligned}
\end{equation}

where $(x,t) \in [-5,5] \times [0,6]$, and the exact solution:

\begin{equation}\label{eq:16}
    \begin{aligned}
u(x, t)= & \frac{0.5}{\cosh (2(x-\sqrt{3} t))}-\frac{0.5}{\cosh (2(x-10+\sqrt{3} t))} \\
& +\frac{0.5}{\cosh (2(x+\sqrt{3} t))}-\frac{0.5}{\cosh (2(x+10-\sqrt{3} t))} .
    \end{aligned}
\end{equation}

The solution exhibits four distinct wave packets propagating outward with discontinuous transitions(Fig.\ref{fig:wave_true_u}), posing significant challenges for traditional sampling methods to resolve spatially separated features efficiently. Implementation details are as follows:

\begin{itemize}
	\item RL-PINNs perform one round of adaptive sampling (retaining 1706 high-variation points).
	\item Baseline methods (UNIFORM, RAR, RAD) execute five rounds, each adding 400 collocation points (2000 points total).
    \item Post-sampling training employs 25000 iterations, whereas baselines train for 5000 iterations per round..
\end{itemize}

Tab.\ref{tab_wave} summarizes the computational performance. RL-PINNs achieve a relative $L_2$ error of 0.0053, surpassing UNIFORM (0.0423), RAR (0.0339), and RAD (0.0351) by 87.5\%, 84.4\%, and 84.9\%, respectively. 

Fig.\ref{fig:wave_points} illustrates the collocation points selected by each method. RL-PINNs concentrate samples along the wavefronts, effectively capturing propagating discontinuities. In contrast, baselines distribute points uniformly or cluster redundantly in regions with low solution variation.

Fig.\ref{fig:wave_error} compares absolute prediction errors across the domain. RL-PINNs maintain minimal errors 
near wavefronts, while UNIFORM and RAD exhibit significant discrepancies due to undersampling critical regions. RAR partially resolves wavefronts but accumulates errors from redundant sampling in smooth zones.

These results demonstrate RL-PINNs’ robustness in resolving multi-modal, time-dependent PDEs with discontinuous features, achieving superior accuracy without compromising computational efficiency.

\begin{table}[h]
\caption{Results of Wave equation}
\label{tab_wave}
\centering
\begin{tabular}{cccc}  
\toprule
Wave &  Total sampling time & Total PINNs training time & $L_2$ \\  
\midrule
UNIFORM & 9.98e-3 s & 1557.33 s & 0.0423 \\  
RAR & 0.28 s & 1570.68 s & 0.0339 \\  
RAD & 0.31 s & 1566.51 s & 0.0351 \\  
RL-PINNs & 7.01 s & 1512.33 s & 0.0053 \\  
\bottomrule
\end{tabular}
\end{table}

\subsection{ High-Dimensional Poisson Equation}
\ 
\newline

To demonstrate RL-PINNs’ scalability in high-dimensional settings, we evaluate the framework on a 10-dimensional elliptic equation:

\begin{figure}[h]
    \centering
{
    \includegraphics[width=0.4\textwidth]{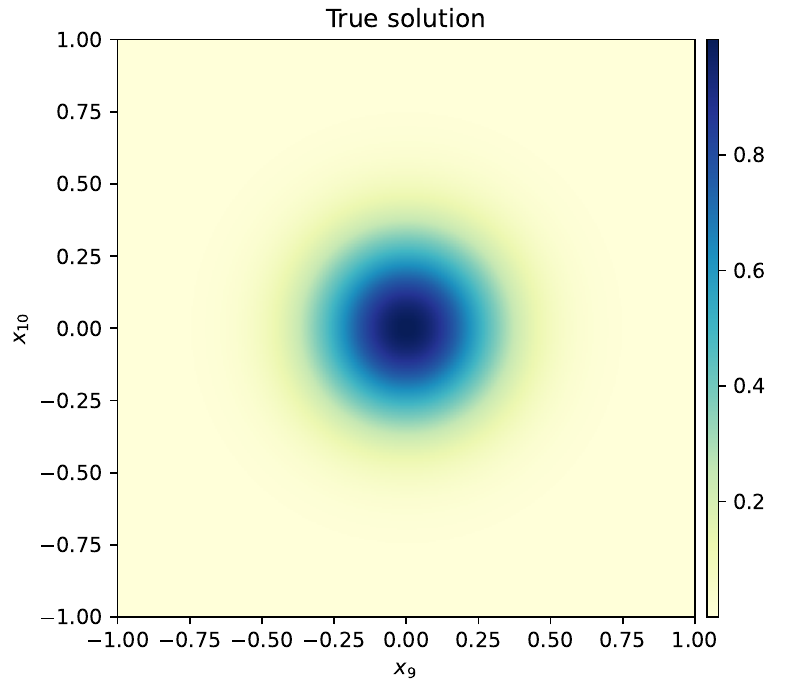}
}
    \caption{\label{fig:high_true_u} High-Dimension: The exact solution.}
\end{figure}

\begin{figure}[h]
    \centering
    \subfigure[UNIFORM]
{
    \includegraphics[width=0.35\textwidth]{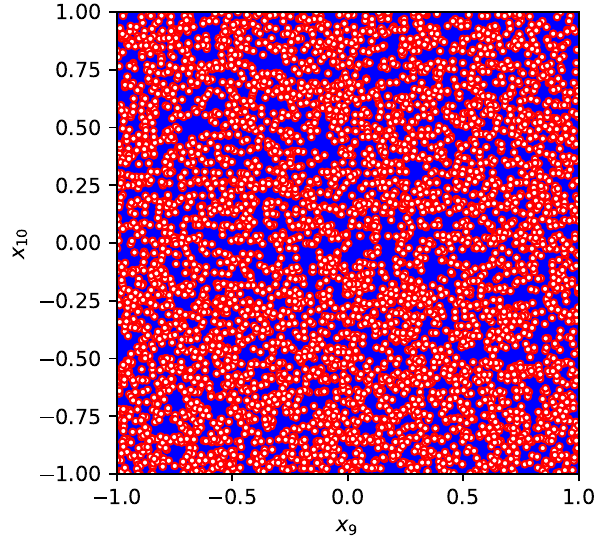}
}
    \subfigure[RAR]
{
    \includegraphics[width=0.35\textwidth]{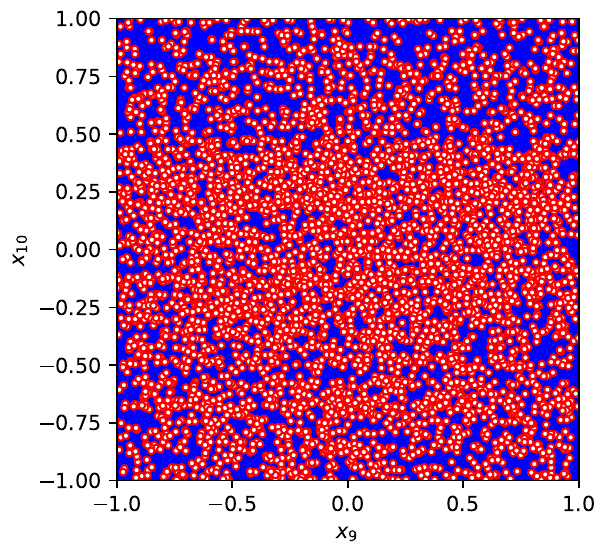}
}
    \subfigure[RAD]
{
    \includegraphics[width=0.35\textwidth]{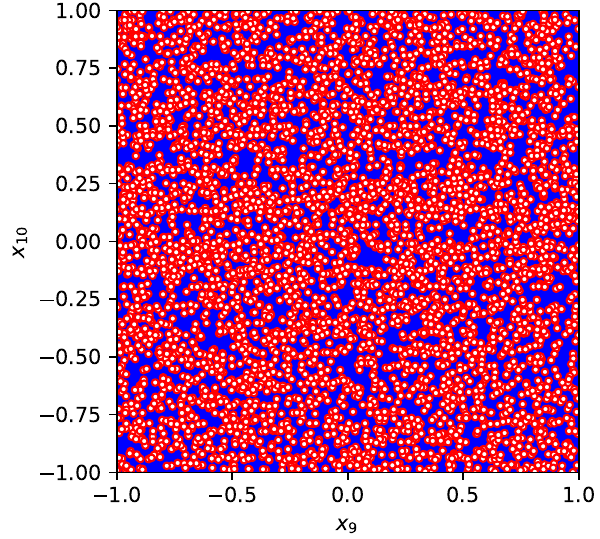}
}
    \subfigure[RL-PINNs]
{
    \includegraphics[width=0.35\textwidth]{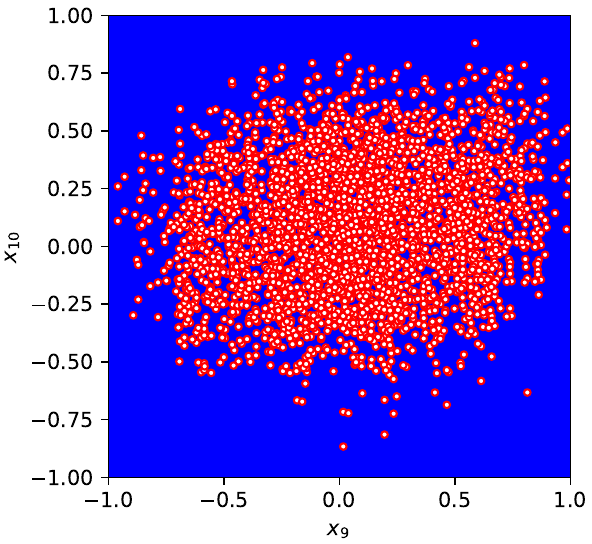}
}
    \caption{\label{fig:high_points} High-Dimension: Cumulative sampling of UNIFORM, RAR, RAD and one-time sampling of RL-PINNs.}
\end{figure}

\begin{figure}[h]
    \centering
{
    \includegraphics[width=0.2\textwidth]{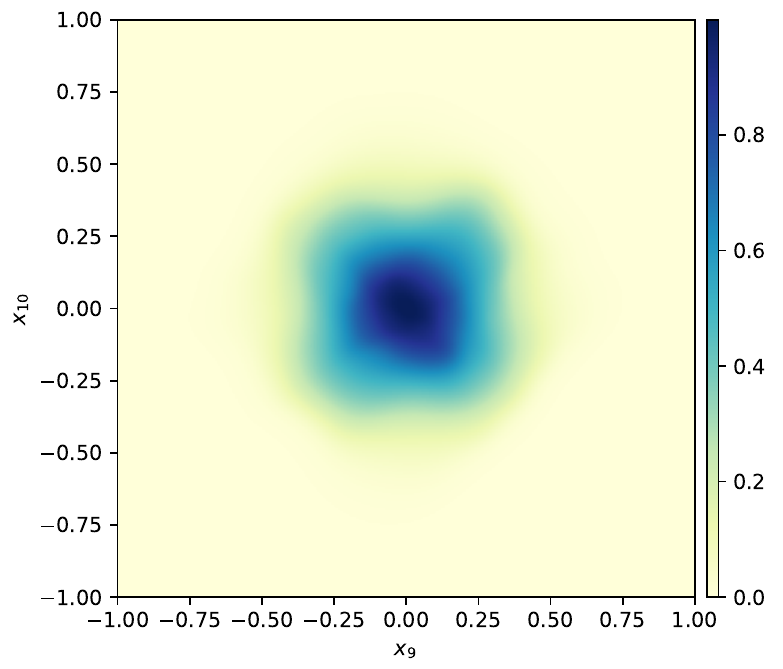}
}
{
    \includegraphics[width=0.2\textwidth]{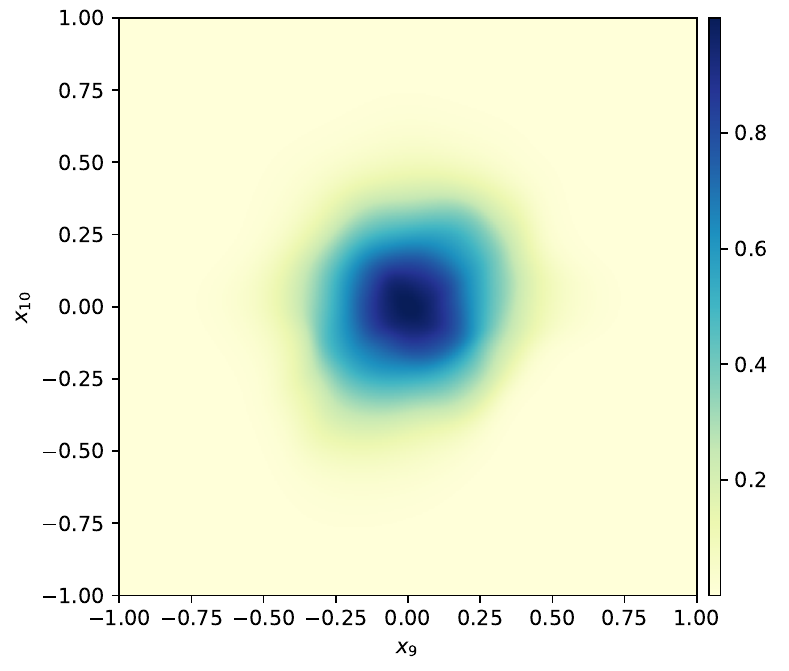}
}
{
    \includegraphics[width=0.2\textwidth]{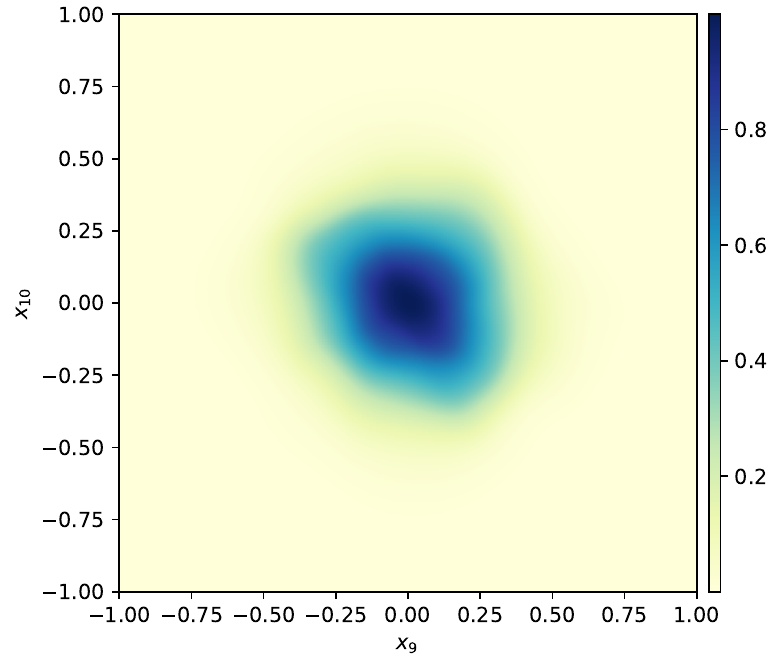}
}
{
    \includegraphics[width=0.2\textwidth]{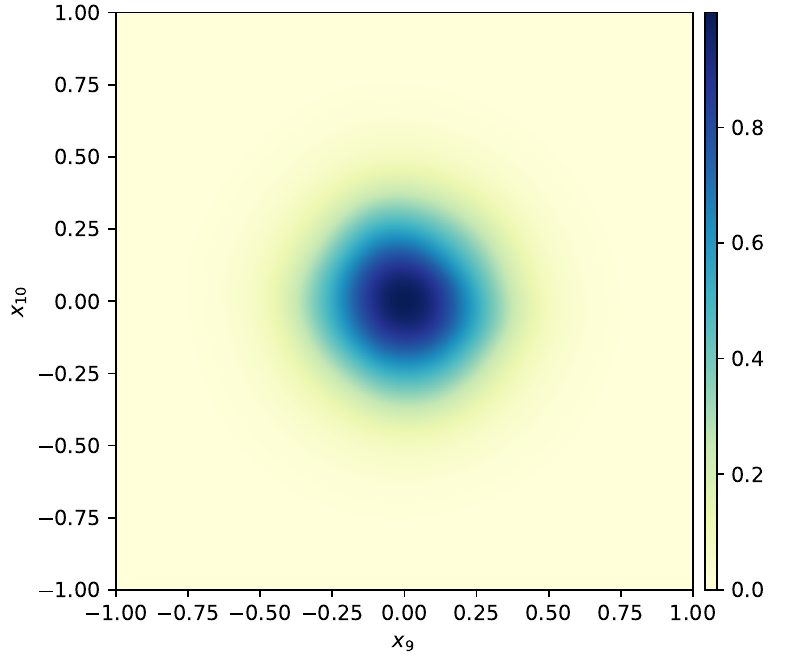}
}
    \subfigure[UNIFORM]
{
    \includegraphics[width=0.2\textwidth]{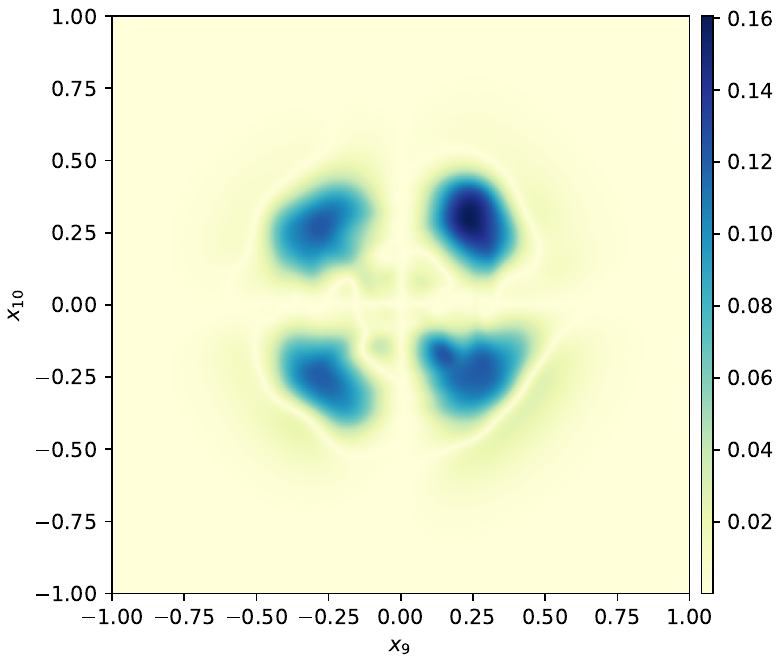}
}
    \subfigure[RAR]
{
    \includegraphics[width=0.2\textwidth]{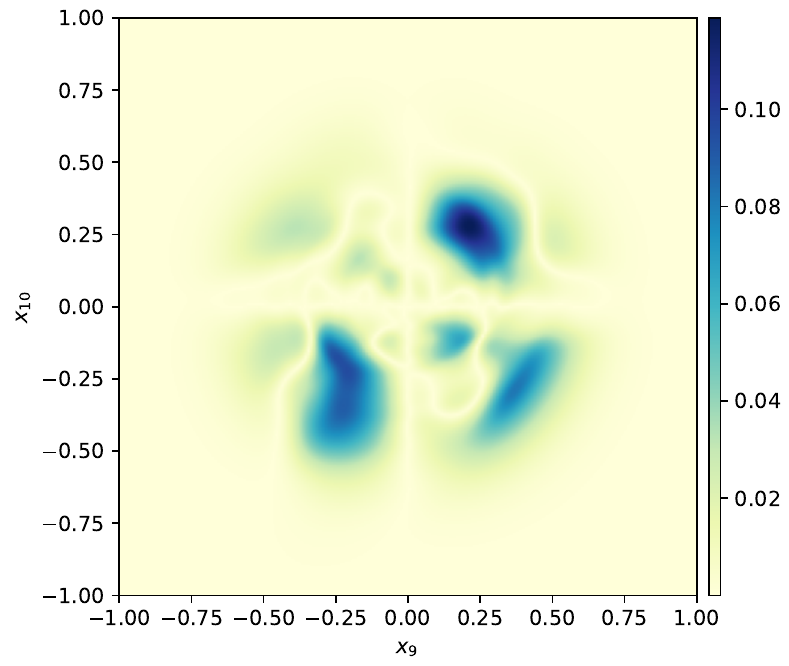}
}
    \subfigure[RAD]
{
    \includegraphics[width=0.2\textwidth]{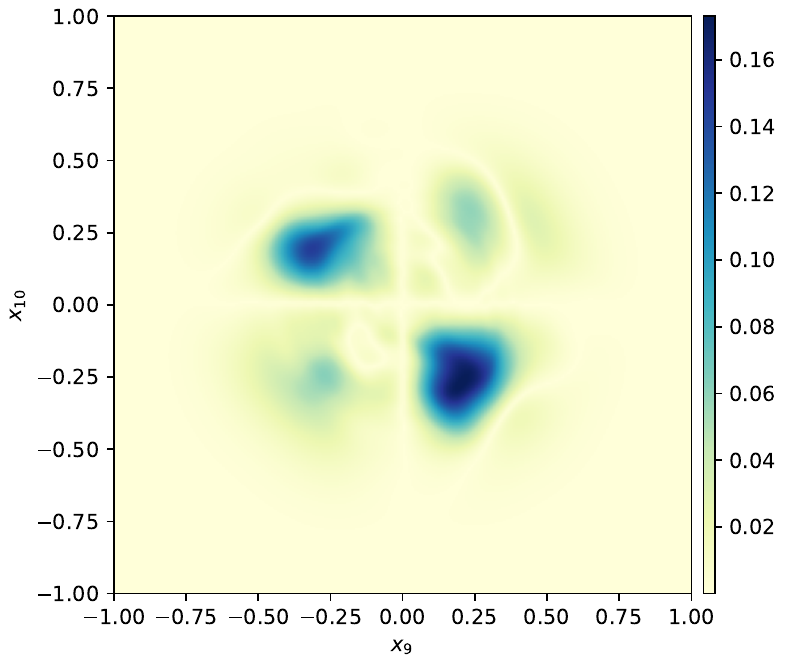}
}
    \subfigure[RL-PINNs]
{
    \includegraphics[width=0.2\textwidth]{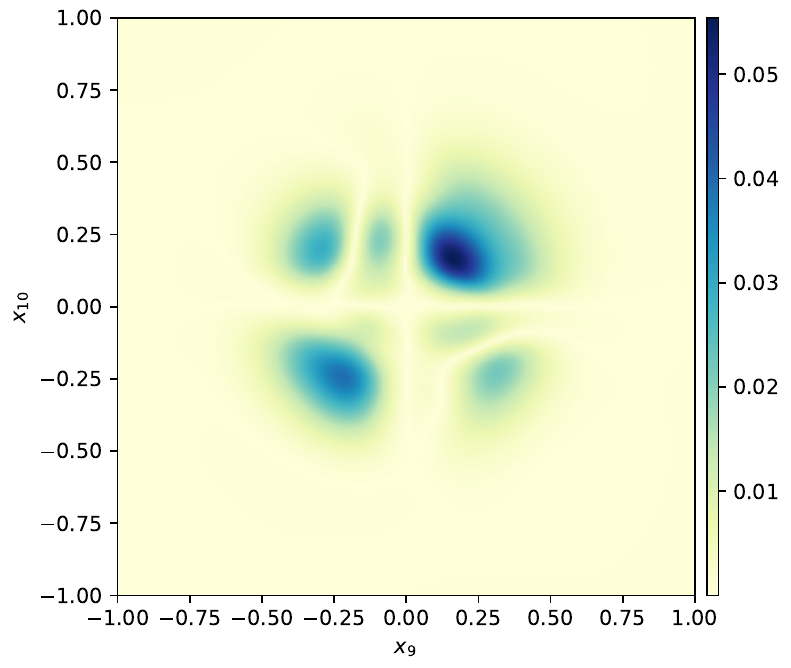}
}
    \caption{\label{fig:high_error}  High-Dimension: Above: The predicted solution. Below: The absolute error of the solution.}
\end{figure}

\begin{equation}\label{eq:17}
    \begin{aligned}
-\Delta u(\mathbf{x})=f(\mathbf{x}), \quad \mathbf{x} \in[-1,1]^{10},
    \end{aligned}
\end{equation}

with the exact solution:

\begin{equation}\label{eq:18}
    \begin{aligned}
u(\mathbf{x})=e^{-10\|\mathbf{x}\|_2^2}.
    \end{aligned}
\end{equation}

where $\|\mathbf{x}\|$ denotes the Euclidean norm. The solution exhibits a sharp exponential decay centered at the origin, posing significant challenges for traditional sampling methods due to the curse of dimensionality. Implementation details are as follows:

\begin{itemize}
	\item RL-PINNs perform one round of adaptive sampling (retaining 3266 high-variation points).
	\item Baseline methods (UNIFORM, RAR, RAD) execute five rounds, each adding 1000 collocation points (5000 points total).
    \item Post-sampling training employs 25000 iterations, whereas baselines train for 5000 iterations per round.
\end{itemize}

Tab.\ref{tab_high} summarizes the results. RL-PINNs achieve a relative $L_2$ error of 0.0394, outperforming UNIFORM (0.1426), RAR (0.0956), and RAD (0.1250) by 72.4\%, 58.8\%, and 68.5\%, respectively. Despite a higher sampling time , the sampling overhead accounts for only 1.27\% of the total runtime.

Fig.\ref{fig:high_points} visualizes the collocation points projected onto a 2D subspace. RL-PINNs concentrate samples near the origin (where the solution decays rapidly), while baselines distribute points uniformly across the domain. This targeted sampling ensures efficient resolution of the high-dimensional solution’s sharp features.

Fig.\ref{fig:high_error} compares absolute errors across test points. RL-PINNs maintain low errors near the origin, whereas UNIFORM and RAD exhibit large discrepancies due to insufficient sampling in critical regions. RAR marginally improves accuracy but fails to match RL-PINNs’ precision due to redundant point selection in smooth regions. These results validate RL-PINNs’ efficacy in high-dimensional PDEs, achieving superior accuracy while mitigating the curse of dimensionality inherent in traditional methods.

\begin{table}[h]
\caption{Results of High-Dimensional Poisson equation}
\label{tab_high}
\centering
\begin{tabular}{cccc}  
\toprule
High-Dimension &  Total sampling time & Total PINNs training time & $L_2$ \\  
\midrule
UNIFORM & 7.04e-3 s & 2811.25 s & 0.1426 \\  
RAR & 0.82 s & 2829.64 s & 0.0956 \\  
RAD & 0.83 s & 2834.97 s & 0.1250 \\  
RL-PINNs & 35.45 s & 2759.82 &0.0394 \\  
\bottomrule
\end{tabular}
\end{table}

\subsection{Biharmonic Equation}
\ 
\newline

We validate RL-PINNs on a fourth-order Biharmonic equation to demonstrate its capability in handling high-order derivative operators. The governing equations are defined as:

\begin{figure}[h]
    \centering
{
    \includegraphics[width=0.4\textwidth]{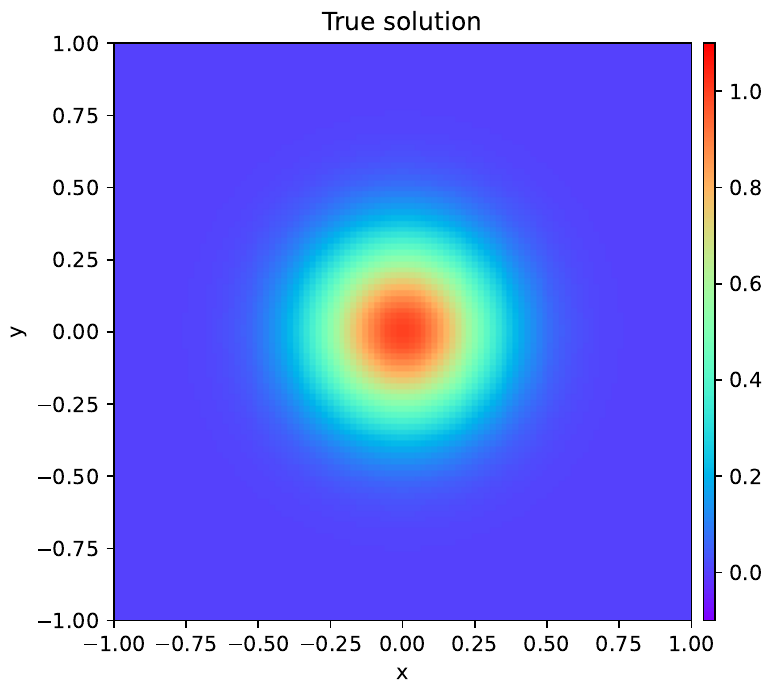}
}
    \caption{\label{fig:order_true_u} High-Order: The exact solution.}
\end{figure}

\begin{figure}[h]
    \centering
    \subfigure[UNIFORM]
{
    \includegraphics[width=0.35\textwidth]{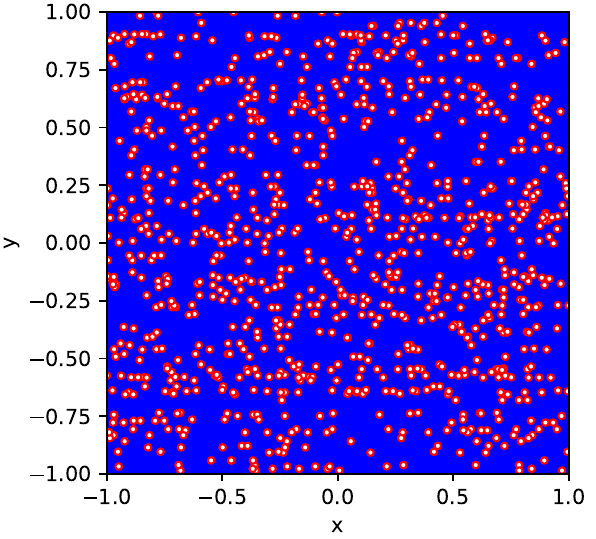}
}
    \subfigure[RAR]
{
    \includegraphics[width=0.35\textwidth]{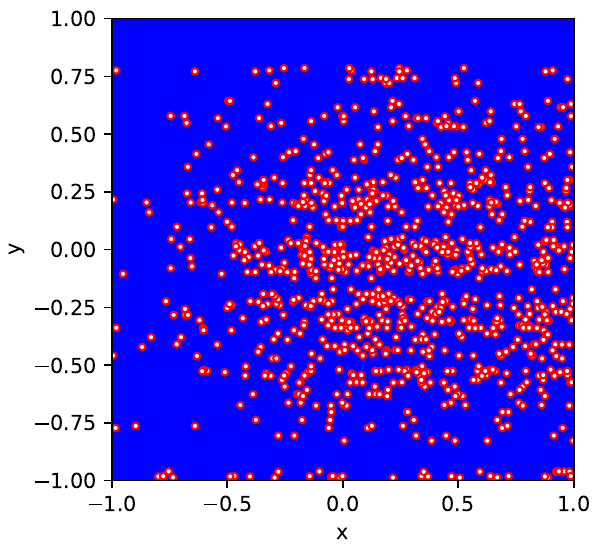}
}
    \subfigure[RAD]
{
    \includegraphics[width=0.35\textwidth]{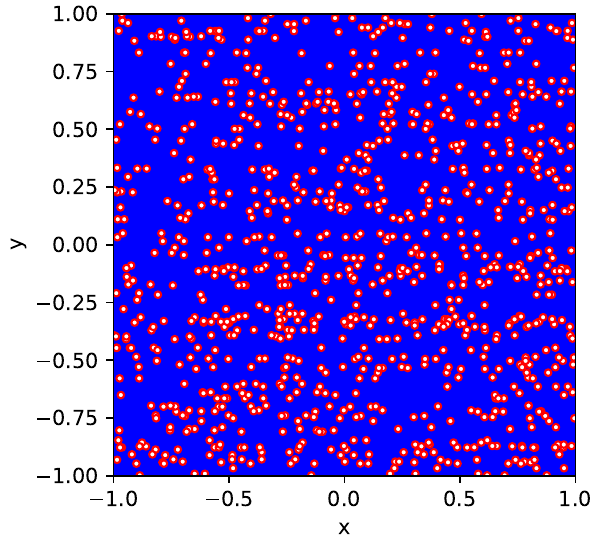}
}
    \subfigure[RL-PINNs]
{
    \includegraphics[width=0.35\textwidth]{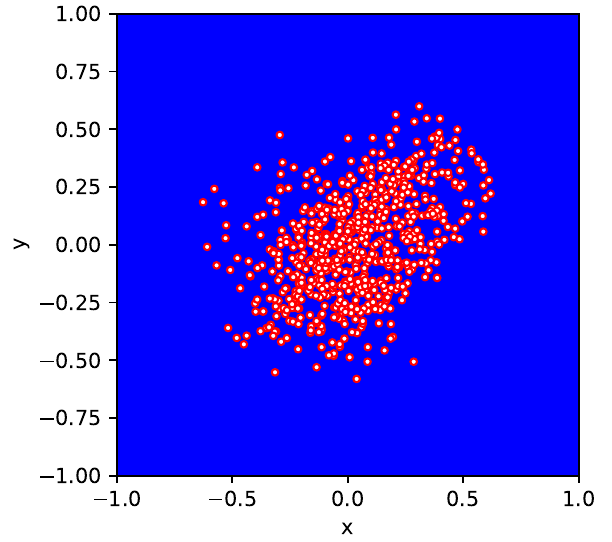}
}
    \caption{\label{fig:order_points} High-Order: Cumulative sampling of UNIFORM, RAR, RAD and one-time sampling of RL-PINNs.}
\end{figure}

\begin{figure}[h]
    \centering
{
    \includegraphics[width=0.2\textwidth]{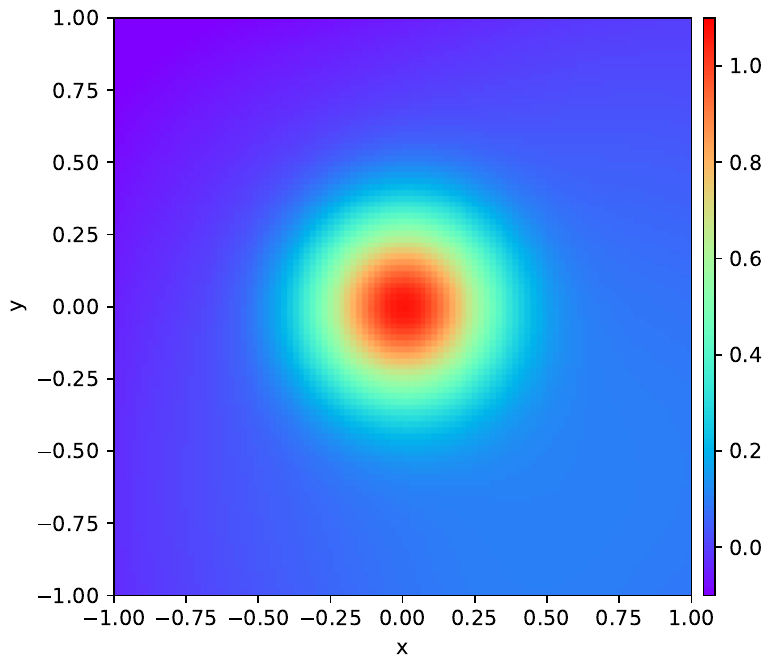}
}
{
    \includegraphics[width=0.2\textwidth]{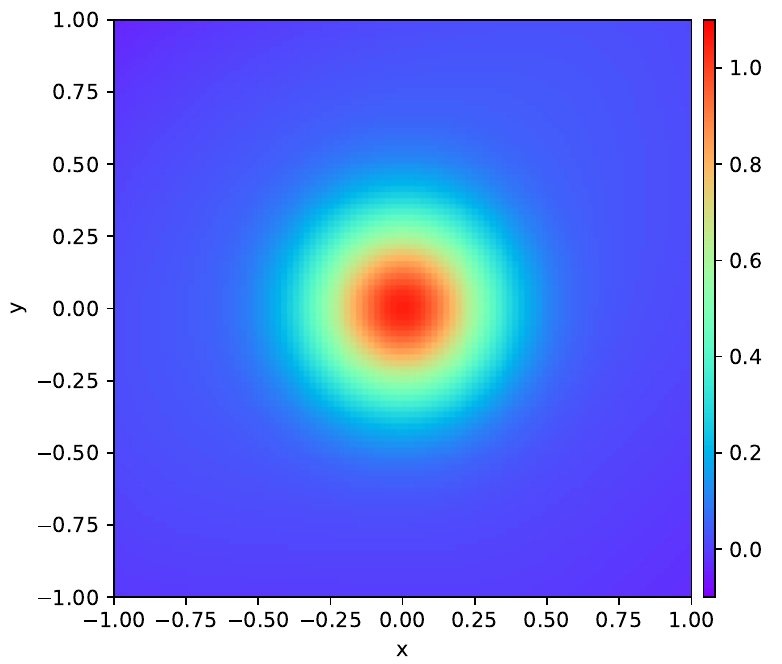}
}
{
    \includegraphics[width=0.2\textwidth]{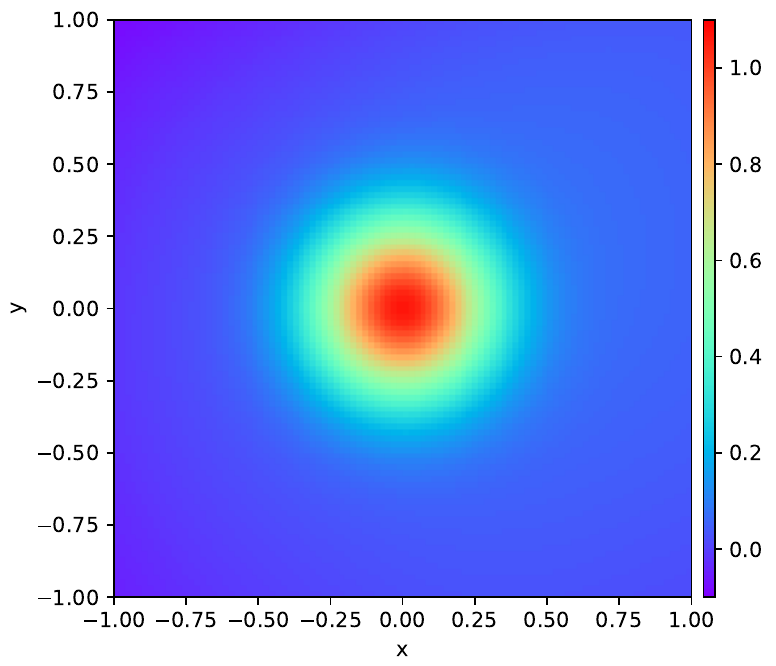}
}
{
    \includegraphics[width=0.2\textwidth]{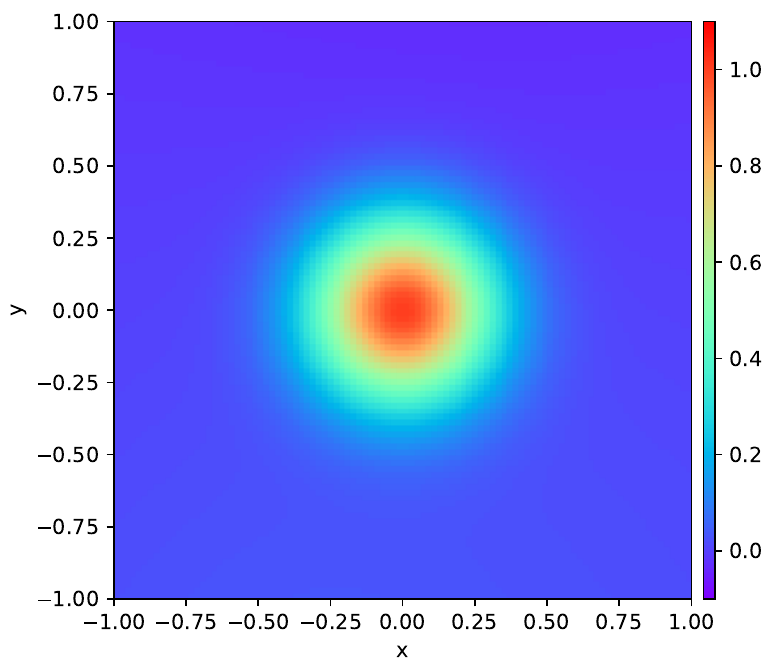}
}
    \subfigure[UNIFORM]
{
    \includegraphics[width=0.2\textwidth]{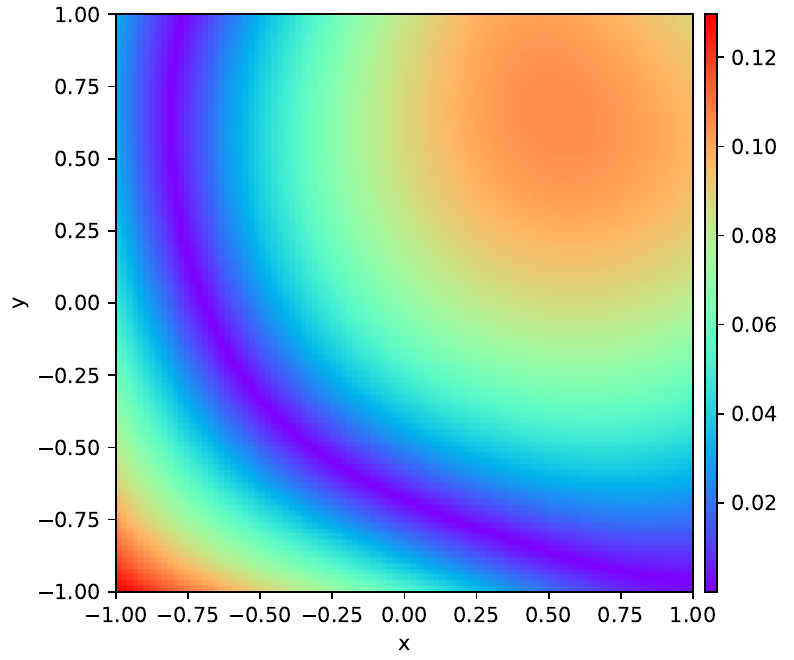}
}
    \subfigure[RAR]
{
    \includegraphics[width=0.2\textwidth]{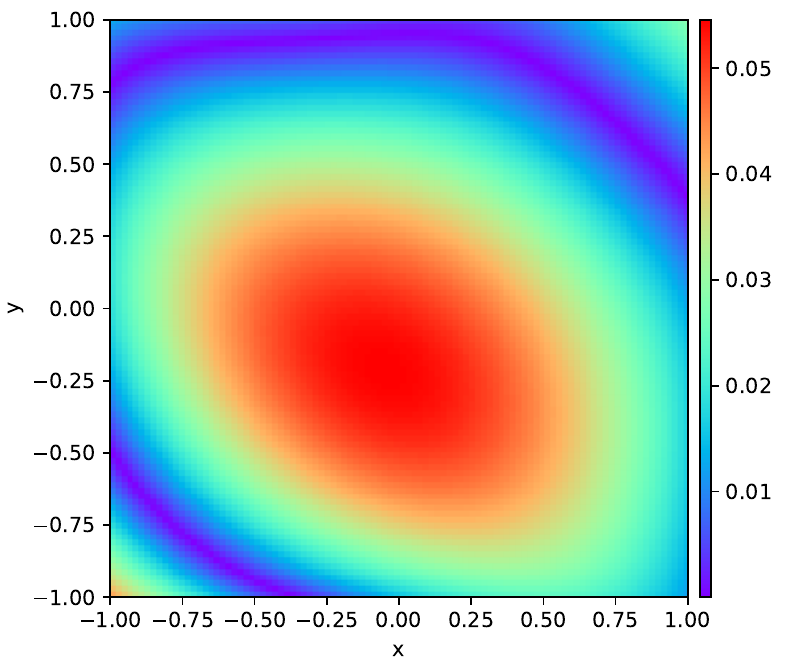}
}
    \subfigure[RAD]
{
    \includegraphics[width=0.2\textwidth]{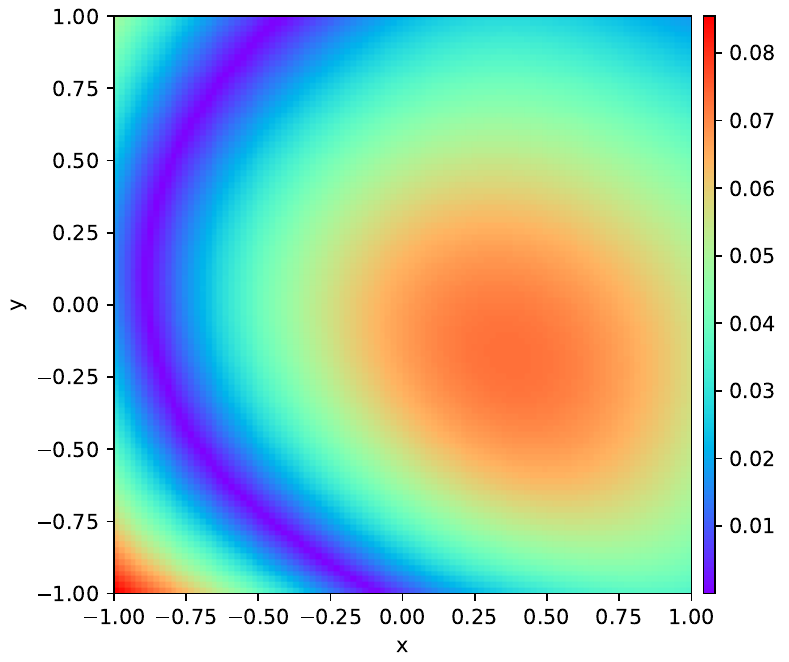}
}
    \subfigure[RL-PINNs]
{
    \includegraphics[width=0.2\textwidth]{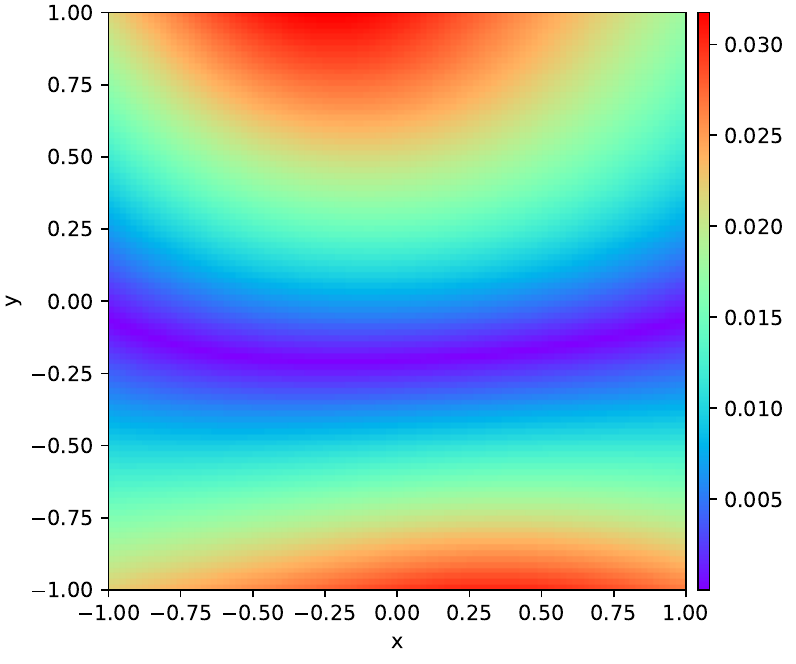}
}
    \caption{\label{fig:order_error}  High-Order: Above: The predicted solution. Below: The absolute error of the solution.}
\end{figure}

\begin{figure}[h]
    \centering
    \subfigure[UNIFORM]
{
    \includegraphics[width=0.2\textwidth]{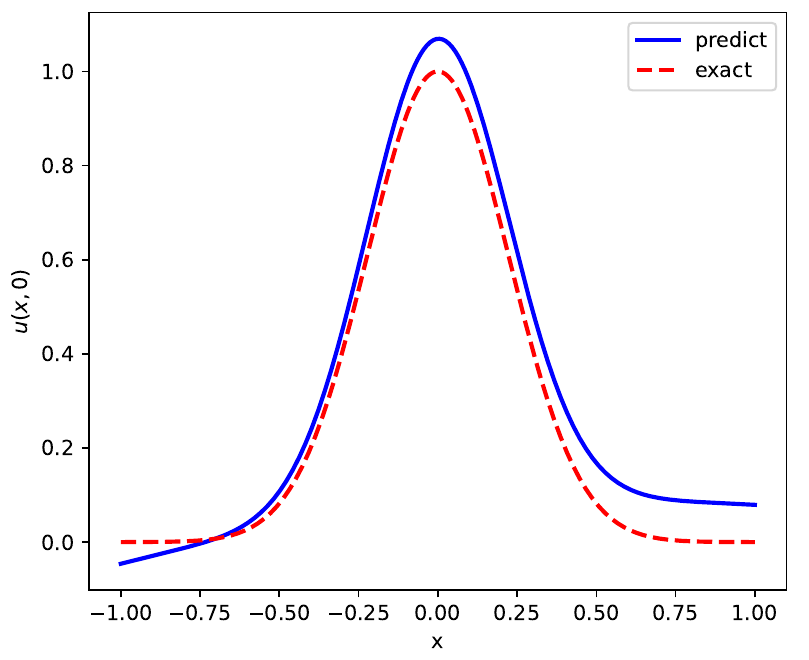}
}
    \subfigure[RAR]
{
    \includegraphics[width=0.2\textwidth]{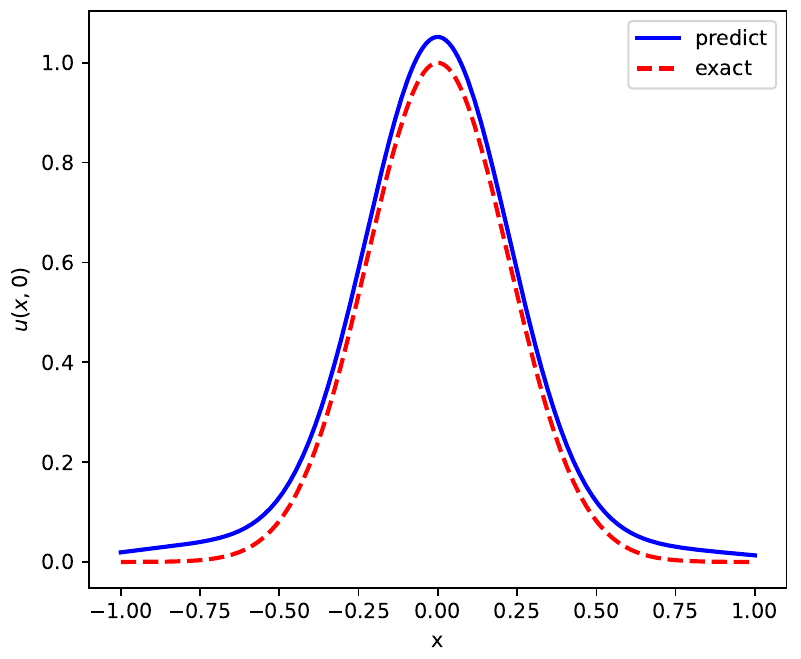}
}
    \subfigure[RAD]
{
    \includegraphics[width=0.2\textwidth]{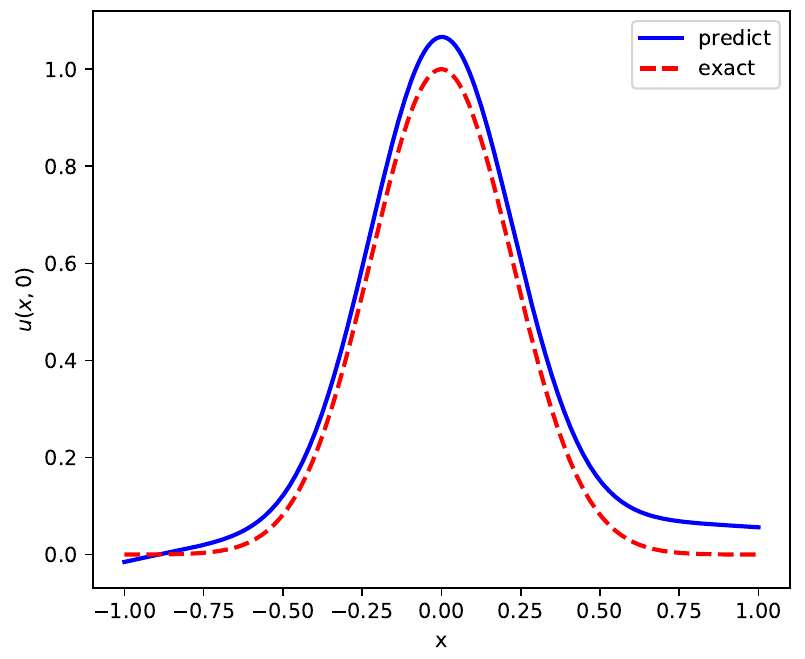}
}
    \subfigure[RL-PINNs]
{
    \includegraphics[width=0.2\textwidth]{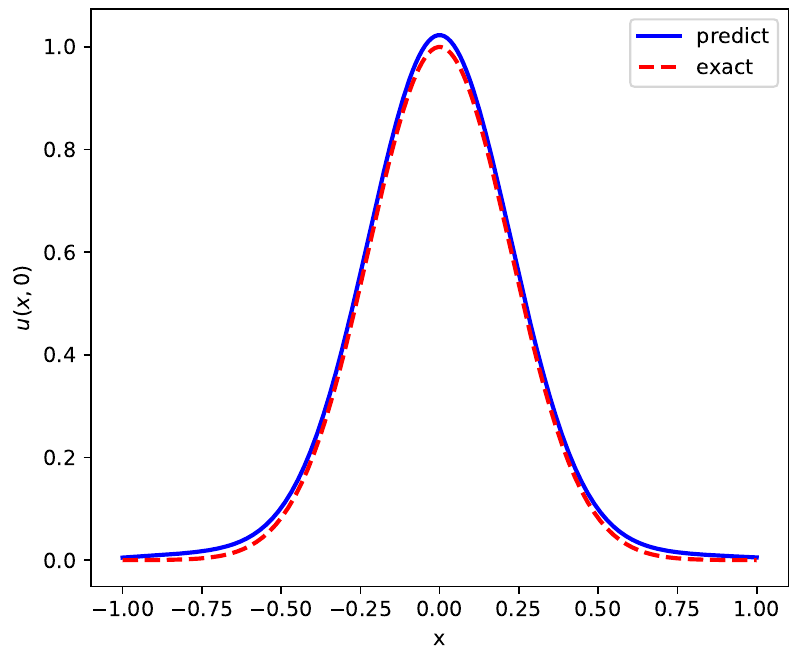}
}
    \caption{\label{fig:order_half}  High-Order: The predicted solution $(x,0)$.}
\end{figure}

\begin{equation}\label{eq:19}
    \begin{aligned}
\left\{\begin{array}{l}
\Delta^2 u(x, y)=f(x, y), \quad  \text { in } \Omega = [-1,1]^2,  \\
u(x, y)=g(x,y), \quad  \text { on }\partial \Omega ,\\
\Delta u(x, y)=h(x,y), \quad  \text { on }\partial \Omega ,
\end{array}\right.
    \end{aligned}
\end{equation}

with the exact solution:

\begin{equation}\label{eq:20}
    \begin{aligned}
u(x,y)=e^{-10(x^2+y^2)},
    \end{aligned}
\end{equation}

where $f(x,y)=\left[160000(x^2+y^2)^2-64000(x^2+y^2)+3200\right] e^{-10(x^2+y^2)}$,  $h(x,y)=\\ \left[400(x^2+y^2)-40\right] e^{-10(x^2+y^2)}$. The solution exhibits a radially symmetric exponential decay with sharp curvature near the origin (Fig.\ref{fig:order_true_u}), challenging conventional adaptive methods due to the fourth-order operator's sensitivity to localized features. Implementation details are as follows:

\begin{itemize}
	\item RL-PINNs perform one round of adaptive sampling (retaining 719 high-variation points).
	\item Baseline methods (UNIFORM, RAR, RAD) execute five rounds, each adding 200 collocation points (1000 points total).
    \item Post-sampling training employs 25000 iterations, whereas baselines train for 5000 iterations per round.
\end{itemize}

Tab.\ref{tab_highorder} summarizes the computational performance. RL-PINNs achieve a relative $L_2$ error of 0.0851, surpassing UNIFORM (0.3265), RAR (0.1611), and RAD (0.2340) by 73.9\%, 47.2\%, and 63.3\%, respectively. 

Fig.\ref{fig:order_points} illustrates the collocation points selected by each method. RL-PINNs densely sample the vicinity of the origin $(0,0)$, where the solution’s fourth-order derivatives dominate. In contrast, baselines distribute points uniformly or redundantly in regions with negligible curvature, failing to resolve the sharp decay effectively.

Fig.\ref{fig:order_error} compares absolute prediction errors. RL-PINNs maintain minimal discrepancies near the origin, whereas UNIFORM and RAD exhibit significant errors due to undersampling critical regions. RAR partially improves accuracy but still underperforms due to redundant points in smooth areas. These results underscore RL-PINNs’ robustness in solving high-order PDEs, achieving superior accuracy without prohibitive computational costs.

\begin{table}[h]
\caption{Results of Biharmonic equation}
\label{tab_highorder}
\centering
\begin{tabular}{cccc}  
\toprule
Case1 &  Total sampling time & Total PINNs training time & $L_2$ \\  
\midrule
UNIFORM & 6.78e-4 s & 3657.54 s & 0.3265 \\  
RAR & 0.52 s & 3655.36 s & 0.1611 \\  
RAD & 0.53 s & 3664.73 s & 0.2340 \\  
RL-PINNs & 4.82 s & 3591.44 s & 0.0851 \\  
\bottomrule
\end{tabular}
\end{table}

\section{Conclusion}

This work proposes RL-PINNs, a novel reinforcement learning-driven adaptive sampling framework for Physics-Informed Neural Networks (PINNs), addressing critical limitations of conventional residual-based adaptive methods.  By reformulating adaptive sampling as a Markov decision process, RL-PINNs enable efficient single-round training while eliminating the computational overhead of gradient-dependent residual evaluations.  Key innovations include: (1)Gradient-Free Reward Design: The use of function variation as a reward metric avoids costly high-order derivative computations, ensuring scalability to high-dimensional and high-order PDEs. (2)Delayed Reward Mechanism: A semi-sparse reward strategy prioritizes long-term training stability, mitigating redundant sampling and oversampling of transient features. (3)Sequential Decision-Making: The RL agent dynamically navigates the solution space to identify critical collocation points, achieving targeted sampling in regions of rapid solution variation. Extensive numerical experiments validate RL-PINNs’ superiority across diverse PDE benchmarks. Potential extensions include integrating RL-PINNs with multi-fidelity models, extending the framework to stochastic PDEs.   Additionally, investigating advanced RL algorithms (e.g., actor-critic methods) could further enhance sampling efficiency and generalization.

\end{document}